\documentclass[10pt,journal,compsoc]{IEEEtran}
\ifCLASSOPTIONcompsoc
  \usepackage[nocompress]{cite}
\else
  \usepackage{cite}
\fi
\ifCLASSINFOpdf
  \usepackage[pdftex]{graphicx}
\else
  \usepackage[dvips]{graphicx}
\fi
\usepackage{amssymb}
\usepackage{amsmath}
\newtheorem{thm}{Theorem}

\usepackage{array}

\ifCLASSOPTIONcompsoc
  \usepackage[caption=false,font=footnotesize,labelfont=sf,textfont=sf]{subfig}
\else
  \usepackage[caption=false,font=footnotesize]{subfig}
\fi
\begin{document}
%

\title{Transferring Knowledge Fragments for Learning Distance Metric from A Heterogeneous Domain}

%
%
%
%

\author{Yong~Luo,
        Yonggang~Wen,~\IEEEmembership{Senior Member,~IEEE,}
        Tongliang~Liu,
        and~Dacheng~Tao,~\IEEEmembership{Fellow,~IEEE}
\IEEEcompsocitemizethanks{
\IEEEcompsocthanksitem Y. Luo and Y. Wen are with the School of Computer Science and Engineering, Nanyang Technological University, Singapore 639798.\protect\\
E-mail: yluo180@gmail.com, ygwen@ntu.edu.sg.
\IEEEcompsocthanksitem T. Liu and D. Tao are with the School of Information Technologies and the Faculty of Engineering and Information Technologies, University of Sydney, Sydney, NSW 2007, Australia.\protect\\
E-mail: tliang.liu@gmail.com, dacheng.tao@sydney.edu.au.
\IEEEcompsocthanksitem \copyright 2018 IEEE. Personal use of this material is permitted. Permission from IEEE must be obtained for all other uses, in any current or future media, including reprinting/republishing this material for advertising or promotional purposes, creating new collective works, for resale or redistribution to servers or lists, or reuse of any copyrighted component of this work in other works.
}
}

\markboth{$>$ \normalsize{TPAMI-2017-04-0282 R}\footnotesize{evision} \normalsize{3} $<$}%
{Shell \MakeLowercase{\textit{et al.}}: Bare Demo of IEEEtran.cls for Computer Society Journals}

\IEEEtitleabstractindextext{%
\begin{abstract}
The goal of transfer learning is to improve the performance of target learning task by leveraging information (or transferring knowledge) from other related tasks. In this paper, we examine the problem of transfer distance metric learning (DML), which usually aims to mitigate the label information deficiency issue in the target DML. Most of the current Transfer DML (TDML) methods are not applicable to the scenario where data are drawn from heterogeneous domains. Some existing heterogeneous transfer learning (HTL) approaches can learn target distance metric by usually transforming the samples of source and target domain into a common subspace. However, these approaches lack flexibility in real-world applications, and the learned transformations are often restricted to be linear. This motivates us to develop a general flexible heterogeneous TDML (HTDML) framework. In particular, any (linear/nonlinear) DML algorithms can be employed to learn the source metric beforehand. Then the pre-learned source metric is represented as a set of knowledge fragments to help target metric learning. We show how generalization error in the target domain could be reduced using the proposed transfer strategy, and develop novel algorithm to learn either linear or nonlinear target metric. Extensive experiments on various applications demonstrate the effectiveness of the proposed method.
\end{abstract}

\begin{IEEEkeywords}
Transfer learning, distance metric learning, heterogeneous domains, knowledge fragments, nonlinear.
\end{IEEEkeywords}}

\maketitle

\IEEEdisplaynontitleabstractindextext

%
\IEEEpeerreviewmaketitle

\IEEEraisesectionheading{\section{Introduction}\label{sec:Introduction}}

%
%
%
%

In the machine learning and pattern recognition applications, we often encounter the label deficiency issue due to the high labeling cost (labor-intensive and expensive). Transfer learning \cite{SJ-Pan-and-Q-Yang-TKDE-2010, H-Azizpour-et-al-TPAMI-2016, WS-Chu-et-al-TPAMI-2017} is able to mitigate this problem in the target learning task or domain by leveraging information from other related source tasks or domains. A typical example is the unconstrained face verification, which is to decide whether two given face images are from the same person or not in wild conditions \cite{JL-Hu-et-al-CVPR-2015}. In a new scenario, we may want to label only a few pairs of face images, and the performance may be very bad. To improve the performance, we can utilize the large amounts of labeled face pairs in some other scenarios. However, the data distribution of two different scenarios can be very different due to the varying lighting and background. Hence the verification model trained in a label-rich scenario may perform unsatisfactorily in the label-scarce one, and transfer learning is helpful in this case. Some other examples include sentiment classification \cite{SJ-Pan-and-Q-Yang-TKDE-2010} and image super-resolution \cite{DX-Dai-et-al-CVPR-2015}.

There are many different types of transfer learning approaches, and we focus on transfer distance metric learning (DML) \cite{Y-Zhang-and-DY-Yeung-TIST-2012, JL-Hu-et-al-CVPR-2015}. DML aims to learn a reliable distance metric to reveal the data relationships, and it is crucial in diverse research areas, ranging from clustering \cite{EP-Xing-et-al-NIPS-2002} and classification \cite{KQ-Weinberger-et-al-NIPS-2005} to kernel machines \cite{ZX-Xu-et-al-arXiv-2013} and ranking \cite{D-Lim-and-G-Lanckriet-ICML-2014}. It usually needs large amount of label information (class labels or similar/dissimilar constraints) to learn reliable distance metric, and DML may fail when the information is insufficient in the target domain. Therefore, it is desirable to leverage information (transfer knowledge) from other related source domains. The similar/dissimilar constraints are weakly-supervised since the exact label for a single sample may be unknown \cite{EP-Xing-et-al-NIPS-2002, A-Bellet-et-al-arXiv-2014}.

The transfer DML (TDML) algorithms can be roughly grouped as homogenous TDML and heterogeneous TDML. In homogeneous TDML, the source and target domain share the same feature representation. This may be not valid in practice. For example, in document categorization, we may want to utilize the abundant labeled English document to help classify the Spanish documents. The document representations of different languages vary since the utilized vocabularies are different. In image annotation and retrieval, it is advantageous to utilize the ``expensive'' (such as deep CNN \cite{K-Chatfield-et-al-arXiv-2014}) or easily interpretable (such as text) feature to guide learning a better representation for the ``cheap'' feature (such as GIST \cite{A-Oliva-and-A-Torralba-IJCV-2001} and LBP \cite{T-Ojala-et-al-TPAMI-2002}) or the one that is harder to interpret (such as most visual features) \cite{GJ-Qi-et-al-SDM-2012, DX-Dai-et-al-CVPR-2015}. The ``expensive'' feature has stronger expressive power (more powerful in describing and capturing intrinsic character of an object and has stronger discriminant ability) than the ``cheap'' feature, but the computational time and/or memory requirement is higher. Extracting ``cheap'' feature instead of ``expensive'' feature can significantly reduce the response time and improve user experience in real-world applications (such as image retrieval), especially on handheld devices. The feature spaces of ``expensive'' and ``cheap'' visual domains are quite different and there is even semantic gap between the text and visual domains.

\begin{figure*}[!t]
\centering
\includegraphics[width=1.6\columnwidth]{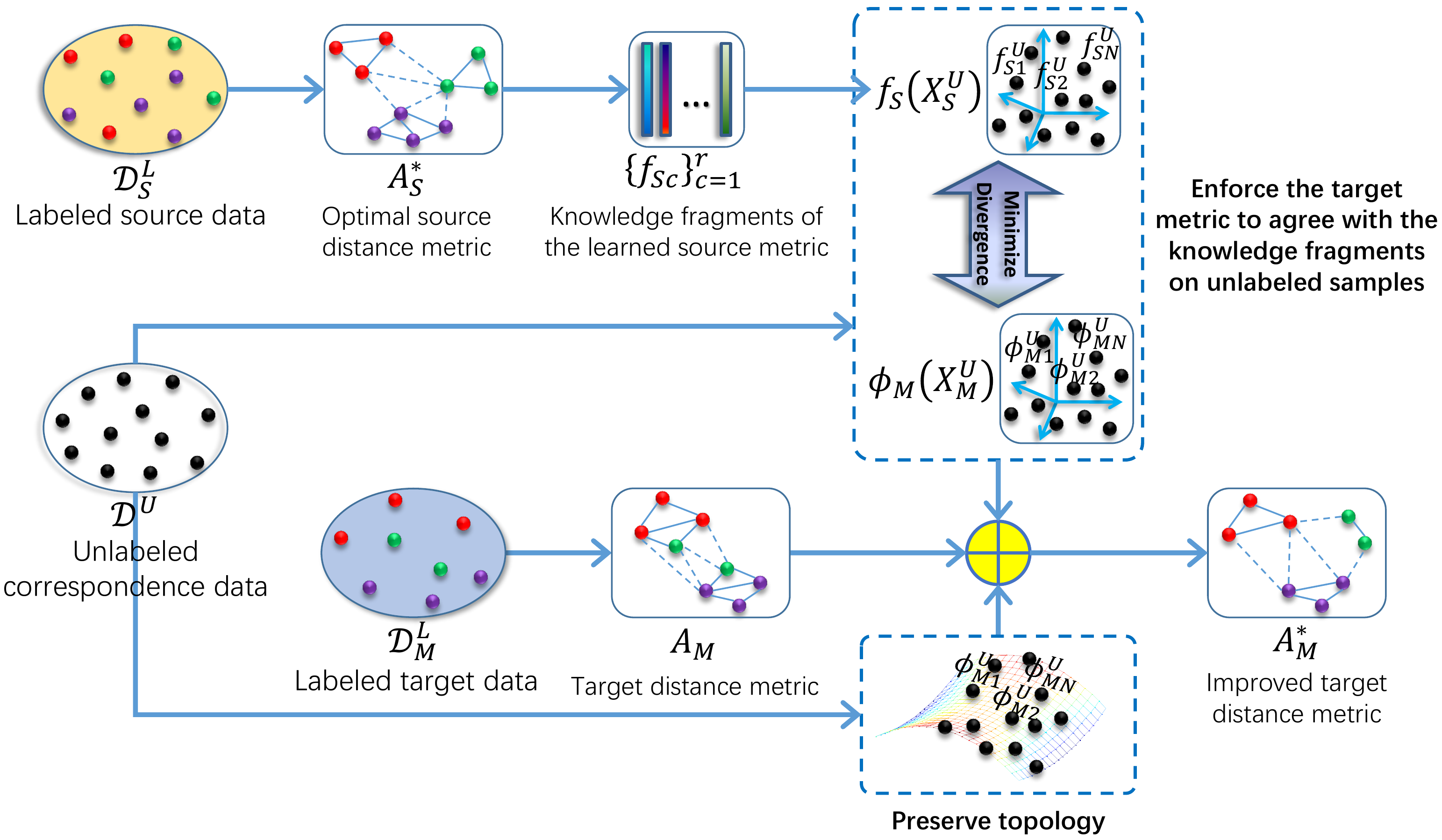}
\caption{Main procedure of the proposed general heterogeneous transfer distance metric learning framework. Samples of the source and target domain lie in different feature spaces. Knowledge fragments $\{f_{Sc}(\cdot)\}$ are extracted from the distance metric learned beforehand in the source domain, and the target metric is formulated as learning a set of mappings $\{\phi_{Mc}(\cdot)\}$. By minimizing the divergence between the unlabeled cross-domain data mapped using $\{f_{Sc}(\cdot)\}$ and $\{\phi_{Mc}(\cdot)\}$, and also preserving the topology in the target domain, we learn improved target distance metric, where the source domain knowledge and unlabeled information is leveraged.} 
\label{fig:System_Diagram}
\end{figure*}

To manage heterogeneous representations, many heterogeneous transfer learning (HTL) \cite{XX-Shi-et-al-ICDM-2010, C-Wang-and-S-Mahadevan-IJCAI-2011, Y-Zhang-and-DY-Yeung-AAAI-2011, GJ-Qi-et-al-SDM-2012} approaches have been proposed in the literature. A frequently utilized strategy in these approaches is to transform the heterogeneous features into a common subspace, where the difference between heterogeneous domains is reduced. Most of the HTL methods are not specially designed for distance metric learning (DML), but we can derive a metric from the transformation learned for each domain. The work of \cite{GJ-Qi-et-al-SDM-2012} is able to learn a target distance function by using large amounts of corresponding data. Although effective in some cases, the current HTL approaches exhibit two main drawbacks: 1) the source and target transformations are learned together. Consequently, they are not feasible when original labeled source domain data are not available (e.g., due to privacy or security reasons). For example, the label information may be private in some applications, such as the user identity information in the user identification application. Besides, learning both the source and target transformations may significantly increase the (time and memory) complexity of the algorithms when the number of samples in the source domain is large. Since our ultimate goal is to learn a good distance metric for the target domain, it is not necessary to include the original labeled source data in the target transformation (metric) learning; 2) the transformations are restricted to be linear and thus the performance may be unsatisfactory in many visual analysis-based applications, since the structures of data distributions are nonlinear for most types of visual features.

To remedy these drawbacks, we develop a general heterogeneous TDML (HTDML) framework inspired by the knowledge fragment transfer strategy \cite{V-Vapnik-and-R-Izmailov-JMLR-2015}. The main procedure is shown in Fig. \ref{fig:System_Diagram}. In particular, the proposed HTDML first learns the source distance metric by applying existing linear or nonlinear metric learning algorithms on given labeled source data. Then we extract some knowledge fragments from the learned metric for transfer. In this paper, we assume there are abundant corresponding unlabeled samples that have feature representations in both of the source and target domains. The metric in the target domain is learned by minimizing the empirical losses w.r.t. the metric and preserving the topology. By further enforcing the metric to agree with the knowledge fragments on the unlabeled samples, we learn an improved target metric, where additional source information contained in the fragments is utilized.

We theoretically illustrate how the source domain could help the target metric learning by examining the generalization error of the proposed HTDML. We develop an algorithm to learn linear metric for the target domain, and extend it to the nonlinear case by incorporating a nonlinear learning technique, the gradient boosting machine \cite{JH-Friedman-AoS-2001, S-Tyree-et-al-WWW-2011, D-Kedem-et-al-NIPS-2012}. The main advantages of the proposed HTDML are: 1) the source knowledge fragments can be learned offline and separately (from the target metric), we do not have to reuse the original labeled source data. Hence the algorithm can be used in the applications where labeled source data are invisible. Although we utilize abundant unlabeled corresponding data to enable knowledge transfer, there is no overlap between the unlabeled data and labeled source data. The unlabeled data do not have label information. Hence, they are much easy to collect, and the privacy of the source domain samples can be well protected. Besides, any (linear or nonlinear) metric learning algorithms can be adopted to learn the source knowledge fragments. Thus, the method is general, flexible, and easy-to-use; 2) nonlinear metric can be learned for the target domain. Hence the proposed method can be widely adopted in many applications, especially the challenging visual-analytic based ones, where the data are often highly nonlinear. We conduct experiments on various popular applications: scene classification, object recognition, image retrieval and face verification. In addition to the Euclidean (EU) and single domain DML baselines, we further compare with several representative heterogeneous transfer learning approaches that could learn distance metric \cite{C-Wang-and-S-Mahadevan-IJCAI-2011, Y-Zhang-and-DY-Yeung-AAAI-2011, GJ-Qi-et-al-SDM-2012, DX-Dai-et-al-CVPR-2015}. The results validate the effectiveness of the proposed HTDML. For example, we have a more than $10\%$ relative improvement compared with all other approaches when limited labeled data are provided.

%
%

\newcommand{\tabincell}[2]{\begin{tabular}{@{}#1@{}}#2\end{tabular}}
\begin{table*}[!t]
\renewcommand{\arraystretch}{1.3}
\caption{Main notations and corresponding definitions (subscript ``$S$'' denotes ``source'' and ``$M$'' signifies ``target'')}
\label{tab:Main_Notations}
\centering
\begin{tabular}{|c||c|}
\hline
Notation & Definition\\
\hline
$(\mathbf{x}_{Si}^1, \mathbf{x}_{Si}^2)$, $(\mathbf{x}_{Mi}^1, \mathbf{x}_{Mi}^2)$ & \tabincell{c}{Feature representations of the $i$-th labeled training pair, i.e., subscripts ``$Si$'' and ``$Mi$'' \\denote the $i$-th labeled training pair in the source and target domain respectively.}\\
\hline
$y_{Si}$, $y_{Mi}$ & Similar/dissimilar constraint for the $i$-th labeled training pair.\\
\hline
$d_S$, $d_M$ & Feature dimensionality.\\
\hline
$N_S$, $N_M$ & Number of labeled training pairs.\\
\hline
$(\mathbf{x}_{Sn}^U, \mathbf{x}_{Mn}^U)$ & \tabincell{c}{Feature representations of the $n$-th unlabeled training sample in both the source and target domain. \\ The superscript ``$U$'' signifies ``unlabeled''.}\\
\hline
$N^U$ & Number of unlabeled cross-domain correspondences.\\
\hline
$A_S$, $A_M$ & Parameter matrix of the distance metric.\\
\hline
$\phi_S$, $\phi_M$ & Feature mapping, which consists of a set of mapping functions and the $c$-th function is denoted as $\phi_{Sc}$ and $\phi_{Mc}$.\\
\hline
$\tilde{p}$, $p_S$ & Fundamental element.\\
\hline
$\tilde{\kappa}(\cdot, \cdot)$, $\kappa_S(\cdot, \cdot)$ & Pre-defined kernel function.\\
\hline
$f_S$ & Knowledge fragments. The $c$-th fragment is given by $f_{Sc}$, which is also a mapping function.\\
\hline
\end{tabular}
\end{table*}

\section{Related work}\label{sec:Related_Work}

Our method is mainly related to distance metric learning and heterogeneous knowledge transfer.

\subsection{Distance metric learning}
How to determine the distance (or similarity) between data is a fundamental problem in machine learning and pattern recognition. Most subsequent learning or recognition tasks will become trivial if perfect distances are obtained. Distance metric learning (DML) aims to solve this problem by learning a distance function to appropriately reflect the data relationships. We can roughly group the DML approaches as linear and nonlinear \cite{B-Kulis-FTML-2012}. Some representative linear DML algorithms include the first Mahalanobis metric learning work presented in \cite{EP-Xing-et-al-NIPS-2002}, Neighborhood Component Analysis (NCA) \cite{J-Goldberger-et-al-NIPS-2004}, Large-Margin Nearest Neighbors (LMNN) \cite{KQ-Weinberger-et-al-NIPS-2005}, Information Theoretic Metric Learning (ITML) \cite{JV-Davis-et-al-ICML-2007}, Regularized Distance Metric Learning (RDML) \cite{R-Jin-et-al-NIPS-2009}, etc. In some cases, the linear Mahalanobis DML algorithms can be extended to learn nonlinear metrics by simply applying the kernel PCA (KPCA) trick \cite{R-Chatpatanasiri-et-al-NEUCOM-2010}. Alternatively, one can learn nonlinear metric by utilizing some nonlinear function learning techniques (such as neural networks \cite{S-Chopra-et-al-CVPR-2005} and gradient boosting regression tree (GBRT) \cite{D-Kedem-et-al-NIPS-2012}), or adopting some nonlinear distance measure (such as Hamming distance \cite{M-Norouzi-et-al-NIPS-2012}).

\subsection{Heterogeneous transfer learning}
Most current heterogeneous transfer learning (HTL) methods handle heterogeneous domains by finding common subspace for them \cite{XX-Shi-et-al-ICDM-2010, C-Wang-and-S-Mahadevan-IJCAI-2011, Y-Zhang-and-DY-Yeung-AAAI-2011}, or learning asymmetric transformation between them \cite{W-Li-et-al-TPAMI-2014, JT-Zhou-et-al-AISTATS-2014}. Although they are not designed for DML, a distance metric can be derived for the target domain if transformations are learned separately for different domains \cite{C-Wang-and-S-Mahadevan-IJCAI-2011, Y-Zhang-and-DY-Yeung-AAAI-2011}. The HTL approach presented in \cite{GJ-Qi-et-al-SDM-2012} is specially designed for DML, but it still relies on learning a common subspace for the source and target domains. Consequently, all these HTL methods lack flexibility in real-world applications, as mentioned in Section \ref{sec:Introduction}.

There exists a recent work of metric imitation \cite{DX-Dai-et-al-CVPR-2015}, which learns an improved distance metric for the target features by utilizing them to approximate the manifold of source domain. It is more flexible than the other HTL approaches since the graph of the source domain can be computed offline, but the solution is obtained by performing eigenvalue decomposition on a large size graph matrix. Hence it is not efficient when the number of unlabeled sample pairs is large. Besides, it suffers from the same limitation of learning only linear transformation as all the other existed HTL approaches.

To overcome these limitations, we develop a general, flexible, and easy-to-use HTDML framework. The knowledge transfer idea is borrowed from \cite{V-Vapnik-and-R-Izmailov-JMLR-2015}, which is designed for privileged information transfer (PIT). The ``privileged information'' is a concept introduced in \cite{V-Vapnik-and-A-Vashist-NN-2009}, where a new learning paradigm termed learning using privileged information (LUPI) is proposed. LUPI assumes that there is additional privileged information for each sample at the training stage (but not at the test stage). The privileged information refers to some ``intelligent'' information provided by ``Intelligent Teacher''. The privileged information can be provided in various ways, such as explanations and comments for the training samples. In \cite{V-Vapnik-and-R-Izmailov-JMLR-2015}, the privileged information is regarded as additional features, which are used to correct the decision function at the training stage. The authors describe a mechanism of privileged information transfer (PIT) under the theme of support vector machine (SVM) in \cite{V-Vapnik-and-R-Izmailov-JMLR-2015}. Motivated by this idea, we propose a heterogeneous transfer distance metric learning framework, which utilizes relatively good distance metric obtained from the source domain to improve the metric learning in the target domain, where either the utilized feature is ``cheap'' or the label information is limited. The privileged information in \cite{V-Vapnik-and-A-Vashist-NN-2009} is similar to the source domain feature in this paper. Although the features utilized in the privileged space and decision space can be different in PIT, there are significant differences between PIT and the proposed source distance metric transfer: 1) PIT needs abundant data with class labels in the decision space (which is similar to the target domain in our method), and each data is associated with privileged information. That is, there are abundant corresponding data with class labels in PIT. Whereas in our method, the data in the target domain are weakly-supervised and scarce. The weakly-supervised data in the source domain can also be scarce if the utilized features are powerful. The labeled source and target data do not need to have correspondences, and we only require some unlabeled corresponding data; 2) PIT focuses on transferring the definitions of classes, instead of the distance metric in our method. It may need some effort to make the PIT model appropriate for other applications, such as clustering and retrieval, while our learned distance metric can be directly used in different applications.

%
%

\section{Heterogeneous transfer distance metric learning}\label{sec:HTDML_KFT}
Problem setting: we suppose the training set with (weakly-supervised) label information for the target domain is given by $\mathcal{D}_M^L = \{ \mathbf{x}_{Mi}^1, \mathbf{x}_{Mi}^2, y_{Mi} \}_{i=1}^{N_M}$, where $\mathbf{x}_{Mi}^1, \mathbf{x}_{Mi}^2 \in \mathbb{R}^{d_M}$, and $y_{Mi} = \pm1$ indicates that $\mathbf{x}_{Mi}^1$ and $\mathbf{x}_{Mi}^2$ are similar/dissimilar to each other. In the target domain, we have only a few samples with label information, and thus DML may perform poorly. Therefore, we assume there exists a relevant source domain with the training set $\mathcal{D}_S^L = \{ \mathbf{x}_{Si}^1, \mathbf{x}_{Si}^2, y_{Si} \}_{i=1}^{N_S}$, where $\mathbf{x}_{Si}^1, \mathbf{x}_{Si}^2 \in \mathbb{R}^{d_S}$ belong to a different feature space from the target domain samples. In the source domain, either the features are more ``expensive'' than the target domain \cite{DX-Dai-et-al-CVPR-2015}, or the samples (with label information) are abundant, i.e., $N_S \gg N_M$. Therefore, a better distance metric can be obtained in the source domain than the target domain. To enable knowledge transfer, we also assume there are large amounts of unlabeled data that have representations in both the source and target domains, i.e., $\mathcal{D}^U = \{ (\mathbf{x}_{Sn}^U, \mathbf{x}_{Mn}^U ) \}_{n=1}^{N^U}$, where $N^U$ is the number of unlabeled corresponding data (in the source and target domain). Such data are usually easy to collect in practice \cite{GJ-Qi-et-al-SDM-2012}, and there is usually no overlap between the unlabeled and labeled training samples. Our ultimate goal is to learn an appropriate distance metric $A_M$ for the target domain. Definitions of the main notations used throughout this paper are summarized in Table \ref{tab:Main_Notations}.

\subsection{Problem formulation}
We propose a general framework for learning distance metric $A_M$ in the target domain by making use of the information from both target and source domains, as well as the unlabeled data. The framework is motivated by the privileged information transfer strategy presented in \cite{V-Vapnik-and-R-Izmailov-JMLR-2015}, where the knowledge in a privileged space $\tilde{\mathcal{X}}$ is represented as a set of functions $\tilde{\kappa}(\tilde{\mathbf{p}}_c, \tilde{\mathbf{x}}), c = 1, 2 , \ldots, r$. Here, $\tilde{\mathbf{p}}_c$ is called a fundamental element, which is a vector from the space $\tilde{\mathcal{X}}$, and $\tilde{\kappa}(\tilde{\mathbf{p}}_c, \tilde{\mathbf{x}})$ is called a fragment of knowledge with the kernel function $\tilde{\kappa}$. If we choose the quadratic kernel function, i.e., $\tilde{\kappa}(\tilde{\mathbf{x}}_i, \tilde{\mathbf{x}}_j) = \langle \tilde{\mathbf{x}}_i, \tilde{\mathbf{x}}_j \rangle^2$, the fundamental elements can be found exactly by solving an eigenvalue problem \cite{V-Vapnik-and-R-Izmailov-JMLR-2015}. To transfer knowledge from the privileged space to the original decision space $\mathcal{X}$, some functions $\{ \phi_c(\mathbf{x}) \}_{c=1}^r$ are found in space $\mathcal{X}$ to approximate the knowledge fragments $\{ \tilde{\kappa}(\tilde{\mathbf{p}}_c, \tilde{\mathbf{x}}) \}_{c=1}^r$. Here, $\mathbf{x}$ and $\tilde{\mathbf{x}}$ are the representations of a given sample in the original decision space and privileged space respectively.

In \cite{V-Vapnik-and-R-Izmailov-JMLR-2015}, both the findings of fundamental elements and knowledge transfer are under the theme of support vector machines (SVM), where sufficient data with class labels and associated privileged information are provided for training. Our setting is much more challenging in that: 1) we only have weakly-supervised label information for limited data; 2) the weakly labeled data in the target domain are scarce and usually different from the source domain. Consequently, the method presented in \cite{V-Vapnik-and-R-Izmailov-JMLR-2015} is not appropriate for DML, and cannot be used in the heterogeneous transfer setting. This problem can be tackled in the proposed heterogeneous transfer distance metric learning (HTDML) framework.

The framework is based on a generalized notion of the Mahalanobis distance \cite{B-Kulis-FTML-2012}. In the literature of distance metric learning (DML), most methods focus on learning the Mahalanobis distance, which is often denoted as
\begin{equation}
\label{eq:Mahalanobis_Distance}
dst_A(\mathbf{x}_i^1, \mathbf{x}_i^2) = (\mathbf{x}_i^1 - \mathbf{x}_i^2)^T A (\mathbf{x}_i^1 - \mathbf{x}_i^2),
\end{equation}
where $A$ is a positive semi-definite matrix and can be factorized as $A = U U^T$. By applying some simple algebraic manipulations, we have $dst_A(\mathbf{x}_i^1, \mathbf{x}_i^2) = \|U \mathbf{x}_i^1 - U \mathbf{x}_i^2 \|_2^2$. In order to take the structure of data distribution into consideration, we propose to conduct DML in the feature space determined by a mapping $\psi$, i.e., $dst_A(\mathbf{x}_i^1, \mathbf{x}_i^2) = \| U \psi(\mathbf{x}_i^1) - U \psi(\mathbf{x}_i^2) \|_2^2$. Then the distance can be further denoted as
\begin{equation}
\label{eq:Generalized_Mahalanobis_Distance}
dst_\phi(\mathbf{x}_i^1, \mathbf{x}_i^2) = \| \phi(\mathbf{x}_i^1) - \phi(\mathbf{x}_i^2) \|_2^2,
\end{equation}
where $\phi(\cdot) = U \psi(\cdot)$ is an integrated mapping function. It is unspecified and can be either linear or nonlinear. Then the learning of the target metric $A_M$ is reformulated as learning the mapping $\phi_M$, and the general formulation of the proposed HTDML for learning $\phi_M$ is given by
\begin{equation}
\label{eq:HTDML_KFT_General}
\begin{split}
\mathop{\arg \min}_{\phi_M} \epsilon(\phi_M) = & E(\phi_M; \{\mathbf{x}_{Mi}^1, \mathbf{x}_{Mi}^2, y_{Mi}\}) \\
& + \gamma R(\{\phi_{Mc}(\cdot)\}, \{f_{Sc}(\cdot)\}; \{\mathbf{x}_{Mn}^U, \mathbf{x}_{Mn}^U\}) \\
& + \gamma_I R_I(\phi_M; \{\mathbf{x}_{Mn}^U\}),
\end{split}
\end{equation}
where $E(\phi_M) = \frac{1}{N_M} \sum_i L(\phi_M; \mathbf{x}_{Mi}^1, \mathbf{x}_{Mi}^2, y_{Mi})$ is the empirical loss w.r.t. $\phi_M$ in the target domain. We choose $L(\phi_M; \mathbf{x}_{Mi}^1, \mathbf{x}_{Mi}^2, y_{Mi}) = g(y_{Mi} [1 - dst_{\phi_M} (\mathbf{x}_{Mi}^1, \mathbf{x}_{Mi}^2)])$ and adopt the hinge loss for $g$, i.e., $g(z) = \max (0, b-z)$. Here, $b$ is set to be zero, and the distance between any pair of samples is given by (\ref{eq:Generalized_Mahalanobis_Distance}). The regularization term $R(\{ \phi_{Mc}(\cdot) \}, \{ f_{Sc}(\cdot) \})$ is to enforce knowledge transfer from the source domains to the target domain, where $\phi_{Mc}$ is the $c$-th coordinate of the vector-valued mapping function $\phi_M$, and $f_{Sc} (\cdot)$ is the $c$-th fragment of knowledge in source domain. The knowledge transfer is performed by using the mapping functions $\{ \phi_{Mc} (\cdot) \}$ in the target domain to approximate the fragments of knowledge in the source domains $\{ f_{Sc} (\cdot) \}$. In this way, the knowledge in the source domain is incorporated to learn the mapping function $\phi_M$, i.e., the distance metric in the target domain. The regularization term $R_I (\phi_M)$ is to exploit some property or prior in the target domain, and we choose it to be a manifold regularization term, so that the topology of the target domain is preserved in the mapped space. In particular, $R_I (\phi_M) = \frac{1}{(N^U)^2} \sum_{i,j=1}^{N^U} w_{ij} \| \phi_M (\mathbf{x}_{Mi}^U) - \phi_M (\mathbf{x}_{Mj}^U) \|_2^2$, where $w_{ij} = \exp (-{\| \mathbf{x}_{Mi}^U - \mathbf{x}_{Mj}^U \|^2}/{(2\omega^2)})$ is the weight between two neighboring nodes $i$ and $j$ in the data adjacency graph \cite{M-Belkin-et-al-JMLR-2006}. Both $\gamma$ and $\gamma_I$ are non-negative trade-off hyper-parameters, and $\omega$ is the bandwidth hyper-parameter.

The knowledge fragments in the source domain can be found in various ways. If classification labels are available, we can train SVM classifiers and use the obtained support vectors as the fundamental elements, and then construct the knowledge fragments using a pre-defined kernel. However, in DML, we are often only provided with weakly-supervised information, e.g., the similarity/dissimilarity between two samples $\mathbf{x}_i$ and $\mathbf{x}_j$. Therefore, we first learn the metric for source domain using some existing DML algorithm, such as LMNN \cite{KQ-Weinberger-et-al-NIPS-2005}, ITML \cite{JV-Davis-et-al-ICML-2007}, and GB-LMNN \cite{D-Kedem-et-al-NIPS-2012}. The output of a metric learning algorithm can be a distance metric $A_S$ \cite{JV-Davis-et-al-ICML-2007} or feature mapping $\phi_S$ \cite{D-Kedem-et-al-NIPS-2012}. For the distance metric, we decompose it as $A_S = P_S P_S^T$, and the columns $\{ \mathbf{p}_{Sc} \}_{c=1}^r$ of the matrix $P_S$ are adopted as the fundamental elements. Then the source knowledge fragment is given by $f_{Sc} (\cdot) = \kappa_S (\mathbf{p}_{Sc}, \cdot)$, where $\kappa_S$ is a pre-defined kernel in the source domain. For the feature mapping $\phi_S$, the knowledge fragment is directly obtained as $f_{Sc} (\cdot) = \phi_{Sc} (\cdot)$, where $\phi_{Sc}$ is the $c$-th coordinate of $\phi_S$.

Given the pre-trained source knowledge fragment $f_{Sc} (\mathbf{x}_{Sn}^U)$, we propose to minimize the divergence between $\phi_{Mc} (\mathbf{x}_{Mn}^U)$ and $f_{Sc} (\mathbf{x}_{Sn}^U)$, where $\mathbf{x}_{Sn}^U$, $\mathbf{x}_{Mn}^U$ are representations in both of the source and target domains for a given unlabeled sample. To this end, the regularization term in (\ref{eq:HTDML_KFT_General}) can be defined as follows,
\begin{equation}
\label{eq:Divergence_Minimization}
\begin{split}
& R (\{\phi_{Mc} (\cdot)\}, \{f_{Sc} (\cdot)\}) \\
& = \frac{1}{N^U} \sum_{n=1}^{N^U} \left( \sum_{c=1}^r \mathrm{Div}\left(\phi_{Mc}(\mathbf{x}_{Mn}^U), f_{Sc}(\mathbf{x}_{Sn}^U)\right) \right),
\end{split}
\end{equation}
where $r$ is the number of fundamental elements, and $\mathrm{Div}(\cdot, \cdot)$ is a divergence measure, which can be the absolute difference, least squares error, etc.

In this paper, we adopt the absolute difference to suppress the effect of outliers. This leads to the following compact regularization term:
\begin{equation}
\label{eq:Compact_Divergence_Minimization}
R (\Phi_M, K_S) = \frac{1}{N^U} |\Phi_M - F_S|,
\end{equation}
where
\begin{equation}
\notag
\Phi_M = \left[
\begin{array}{ccc}
  \phi_{M1}(\mathbf{x}_{M1}^U) & \cdots & \phi_{M1}(\mathbf{x}_{MN^U}^U) \\
  \vdots & \ddots & \vdots \\
  \phi_{Mr}(\mathbf{x}_{M1}^U) & \cdots & \phi_{Mr}(\mathbf{x}_{MN^U}^U)
\end{array}
\right]
\end{equation}
is the mapped matrix of the unlabeled data in the target domain, and $F_S$ is a knowledge fragment matrix represented by the unlabeled data in the source domain, i.e.,
\begin{equation}
\notag
F_S = \left[
\begin{array}{ccc}
  f_{S1}(\mathbf{x}_{S1}^U) & \cdots & f_{S1}(\mathbf{x}_{SN^U}^U) \\
  \vdots & \ddots & \vdots \\
  f_{Sr}(\mathbf{x}_{S1}^U) & \cdots & f_{Sr}(\mathbf{x}_{SN^U}^U)
\end{array}
\right]
\end{equation}
with each $f_{Sc}(\mathbf{x}_{Sn}^U) = \kappa_S (\mathbf{p}_{Sc}, \mathbf{x}_{Sn}^U)$ or $f_{Sc}(\mathbf{x}_{Sn}^U) = \phi_{Sc}(\mathbf{x}_{Sn}^U)$. Here, $|A| = \sum_i \sum_j |A_{ij}|$ is the sum of all the elements' absolute values for the matrix $A$. By substituting (\ref{eq:Compact_Divergence_Minimization}) into (\ref{eq:HTDML_KFT_General}), we obtain the following specific optimization problem for HTDML:
\begin{equation}
\label{eq:HTDML_KFT_Specific}
\begin{split}
& \mathop{\arg \min}_{\phi_M} \epsilon(\phi_M) \\
& = \frac{1}{N_M} \sum_i g\left( y_{Mi} [1 - \| \phi_M(\mathbf{x}_{Mi}^1) - \phi_M(\mathbf{x}_{Mi}^2) \|_2^2] \right) \\
& \ \ \ \ + \frac{\gamma}{N^U} |\Phi_M - F_S| \\
& \ \ \ \ + \frac{\gamma_I}{(N^U)^2} \sum_{i,j} w_{ij} \| \phi_M(\mathbf{x}_{Mi}^U) - \phi_M(\mathbf{x}_{Mj}^U) \|_2^2.
\end{split}
\end{equation}
We will show how the first regularization term could reduce the generalization error in the target domain by leveraging knowledge from the source domain (with different representation).

\subsection{Theoretical analysis}
In this section, we theoretically illustrate how the proposed model enable the distance metric information to be transferred between heterogeneous domains. The main result is presented in Theorem \ref{thm:Gen_Error_Bound_Target}, which provides a performance guarantee and insights to understand the proposed method. Before showing the main result, we introduce a generalization error for metric learning.

According to (\ref{eq:HTDML_KFT_General}), the objective of HTDML is to learn
\begin{equation}
\label{eq:HTDML_OBJ}
\begin{split}
\hat{\phi}_M = \mathop{\arg \min}_{\phi_M} \epsilon(\phi_M) = & E(\phi_M) + \gamma R(\{\phi_{Mc}(\cdot)\}, \{f_{Sc}(\cdot)\}) \\
& + \gamma_I R_I(\phi_M).
\end{split}
\end{equation}
Ideally, we may prefer an $\phi_M^\ast$ that minimizes the expectation of $\epsilon(\phi_M)$:
\begin{equation}
\label{eq:Exp_Min}
\phi_M^\ast = \mathop{\arg \min}_{\phi_M} \mathbb{E}_{(\mathbf{x}_M^1, \mathbf{x}_M^2, y_M)} \epsilon(\phi_M),
\end{equation}
where the optimal $\phi_M^\ast$ is obtained by fully exploiting all the possible weakly-supervised data in the target domain. Such solution is desired but impossible to obtain since we do not know the distribution of the target domain data. The goodness of the learned $\hat{\phi}_M$ using (\ref{eq:HTDML_KFT_General}) can be evaluated by estimating the difference between $\hat{\phi}_M$ and $\phi_M^\ast$. One way to estimate such difference is to calculate the distance between $\mathbb{E}_{(\mathbf{x}_M^1, \mathbf{x}_M^2, y_M)} \epsilon(\hat{\phi}_M)$ and $\mathbb{E}_{(\mathbf{x}_M^1, \mathbf{x}_M^2, y_M)} \epsilon(\phi_M^\ast)$. If the distance is small, we can say that $\hat{\phi}_M$ and $\phi_M^\ast$ are close to each other.

In particular, we have that
\begin{equation}
\label{eq:Gen_Error}
\begin{split}
& \mathbb{E}_{(\mathbf{x}_M^1, \mathbf{x}_M^2, y_M)} \epsilon(\hat{\phi}_M) - \mathbb{E}_{(\mathbf{x}_M^1, \mathbf{x}_M^2, y_M)} \epsilon(\phi_M^\ast) \\
& = \mathbb{E}_{(\mathbf{x}_M^1, \mathbf{x}_M^2, y_M)} \epsilon(\hat{\phi}_M) - \epsilon(\hat{\phi}_M) + \epsilon(\hat{\phi}_M) \\
& \ \ \ \ - \epsilon(\phi_M^\ast) + \epsilon(\phi_M^\ast) - \mathbb{E}_{(\mathbf{x}_M^1, \mathbf{x}_M^2, y_M)} \epsilon(\phi_M^\ast) \\
& \leq \mathbb{E}_{(\mathbf{x}_M^1, \mathbf{x}_M^2, y_M)} \epsilon(\hat{\phi}_M) - \epsilon(\hat{\phi}_M) \\
& \ \ \ \ + \epsilon(\phi_M^\ast) - \mathbb{E}_{(\mathbf{x}_M^1, \mathbf{x}_M^2, y_M)} \epsilon(\phi_M^\ast) \\
& \leq |\mathbb{E}_{(\mathbf{x}_M^1, \mathbf{x}_M^2, y_M)} \epsilon(\hat{\phi}_M) - \epsilon(\hat{\phi}_M)| \\
& \ \ \ \ + |\epsilon(\phi_M^\ast) - \mathbb{E}_{(\mathbf{x}_M^1, \mathbf{x}_M^2, y_M)} \epsilon(\phi_M^\ast)| \\
& \leq 2 \mathop{\mathrm{sup}}_{\phi_M} |\mathbb{E}_{(\mathbf{x}_M^1, \mathbf{x}_M^2, y_M)} \epsilon(\phi_M) - \epsilon(\phi_M)| \\
& = 2 \mathop{\mathrm{sup}}_{\phi_M} |\mathbb{E}_{(\mathbf{x}_M^1, \mathbf{x}_M^2, y_M)} E(\phi_M) - E(\phi_M)|,
\end{split}
\end{equation}
where the first inequality holds because $\hat{\phi}_M$ is the minimizer of $\epsilon(\phi_M)$ and hence $\epsilon(\hat{\phi}_M) - \epsilon(\phi_M^\ast)$ is non-positive (the solution $\phi_M^\ast$ has the best generalization performance, but not necessary minimize our objective function $\epsilon(\phi_M)$, which is designed for finding an appropriate solution given the limited supervised data). The term $\mathop{\mathrm{sup}}_{\phi_M} |\mathbb{E}_{(\mathbf{x}_M^1, \mathbf{x}_M^2, y_M)} E(\phi_M) - E(\phi_M)|$ is called generalization error for a metric learning algorithm.

We will then bound the generalization error and show that how the proposed model make the error small by exploiting source domain information. By exploiting the bounded difference inequality \cite{S-Boucheron-et-al-Oxford-2013}, Bartlett and Mendelson \cite{PL-Bartlett-and-S-Mendelson-JMLR-2002} have proven the following theorem for the source domain.
\begin{thm}
\label{thm:Gen_Error_Bound_Source}
Assume that the loss function $L(f_S; \mathbf{x}_S^1, \mathbf{x}_S^2, y_S)$ is upper bounded by $B_S$ for the source domain. Then, for any $\zeta>0$, with probability at least $1-\zeta$, we have
\begin{equation}
\label{eq:Gen_Error_Bound_Source}
\begin{split}
& \mathop{\mathrm{sup}}_{f_S} \bigg( \mathbb{E}_{(\mathbf{x}_S^1, \mathbf{x}_S^2, y_S)} L(f_S; \mathbf{x}_S^1, \mathbf{x}_S^2, y_S) - \\ & \ \ \ \ \ \ \ \ \frac{1}{N_S} \sum_{i} L(f_S; \mathbf{x}_{Si}^1, \mathbf{x}_{Si}^2, y_{Si}) \bigg) \\
& \leq 2 \mathfrak{R}(L \circ f_S) + B_S \sqrt{\frac{\log(1/\zeta)}{2 N_S}},
\end{split}
\end{equation}
where
\begin{equation}
\notag
\begin{split}
& \mathfrak{R}(L \circ f_S) \\
& = \mathbb{E}_{\mu, (\mathbf{x}_S^1, \mathbf{x}_S^2, y_S)} \mathop{\mathrm{sup}}_{f_S} \frac{1}{N_S} \sum_i \mu_i L(f_S; \mathbf{x}_{Si}^1, \mathbf{x}_{Si}^2, y_{Si})
\end{split}
\end{equation}
is called the Rademacher complexity and $\{ \mu_i \}$ are Rademacher variables uniformly distributed over the set $\{ -1, 1 \}$.
\end{thm}

We further assume that the loss function $L(f_S; \mathbf{x}_S^1, \mathbf{x}_S^2, y_S)$ is $Z_{f_S}$-Lipschitz, i.e.,
\begin{equation}
\notag
\begin{split}
& |L(f_S; \mathbf{x}_{S1}^1, \mathbf{x}_{S1}^2, y_{S1}) - L(f_S; \mathbf{x}_{S2}^1, \mathbf{x}_{S2}^2, y_{S2})| \\
& \leq Z_{f_S} |\|f_S(\mathbf{x}_{S1}^1) - f_S(\mathbf{x}_{S1}^2)\|_2 - \|f_S(\mathbf{x}_{S2}^1) - f_S(\mathbf{x}_{S2}^2)\|_2|
\end{split}
\end{equation}
for any $(\mathbf{x}_{S1}^1, \mathbf{x}_{S1}^2, y_{S1})$ and $(\mathbf{x}_{S2}^1, \mathbf{x}_{S2}^2, y_{S2})$ in the source domain and some $Z_{f_S} > 0$. By employing the Talagrand contraction Lemma \cite{M-Ledoux-and-M-Talagrand-Springer-2013}, we have
\begin{equation}
\notag
\begin{split}
\mathfrak{R}(L \circ f_S) & = \mathbb{E}_{\mu, (\mathbf{x}_S^1, \mathbf{x}_S^2, y_S)} \mathop{\mathrm{sup}}_{f_S} \frac{1}{N_S} \sum_i \mu_i L(f_S; \mathbf{x}_{Si}^1, \mathbf{x}_{Si}^2, y_{Si}) \\
& \leq Z_{f_S} \mathbb{E}_{\mu, (\mathbf{x}_S^1, \mathbf{x}_S^2)} \frac{1}{N_S} \sum_i \mu_i \| f_S(\mathbf{x}_{Si}^1) - f_S(\mathbf{x}_{Si}^2) \|_2.
\end{split}
\end{equation}
Note that $g(y_S [1 - \|f_S(\mathbf{x}_S^1) - f_S(\mathbf{x}_S^1)\|_2^2 ])$ is $2z$-Lipschitz if we assume that $\| f_S(\mathbf{x}_S) \|_2$ is upper bounded by $z$. Since the metric $f_S$ is learned by using ``expensive'' features or employing a large amount of weakly-supervised data, it is reasonable to assume that it has a good generalization ability and thus $\mathfrak{R}(L \circ f_S)$ is small. The number $Z_{f_S}$ is not small since it is the upper bound of the absolute value of the derivative as mentioned above. Hence, the term $\mathbb{E}_{\mu, (\mathbf{x}_S^1, \mathbf{x}_S^2)} \mathop{\mathrm{sup}}_{f_S} \frac{1}{N_S} \sum_i \mu_i \|f_S(\mathbf{x}_{Si}^1) - f_S(\mathbf{x}_{Si}^2)\|_2$ is small.

Now, we are ready to show the main theoretical result.
\begin{thm}
\label{thm:Gen_Error_Bound_Target}
Assume that the loss function $L(\phi_M; \mathbf{x}_M^1, \mathbf{x}_M^2, y_M)$ is upper bounded by $B_M$ for the target domain and that it is $Z_\phi$-Lipschitz. Then, for any $\zeta > 0$, with probability at least $1 - \zeta$, we have
\begin{equation}
\label{eq:Gen_Error_Bound_Target}
\begin{split}
& \mathop{\mathrm{sup}}_{\phi_M} \bigg( \mathbb{E}_{(\mathbf{x}_M^1, \mathbf{x}_M^2, y_M)} L(\phi_M; \mathbf{x}_M^1, \mathbf{x}_M^2, y_M) - \\ & \ \ \ \ \ \ \ \ \frac{1}{N_M} \sum_{i} L(\phi_M; \mathbf{x}_{Mi}^1, \mathbf{x}_{Mi}^2, y_{Mi}) \bigg) - B_M \sqrt{\frac{\log(1/\zeta)}{2 N_M}} \\
& \leq 2 Z_\phi \mathbb{E}_{\mu, (\mathbf{x}_M^1, \mathbf{x}_M^2)} \frac{1}{N_M} \sum_i \mu_i \| \phi_M(\mathbf{x}_{Mi}^1) - \phi_M(\mathbf{x}_{Mi}^2) \|_2 \\
& \leq 2 Z_\phi \mathbb{E}_{\mu, (\mathbf{x}_M^1, \mathbf{x}_S^1)} \frac{1}{N_M} \sum_i \mu_i \| \phi_M(\mathbf{x}_{Mi}^1) - f_S(\mathbf{x}_{Si}^1) \|_2 \\
& \ \ \ \ + 2 Z_\phi \mathbb{E}_{\mu, (\mathbf{x}_S^1, \mathbf{x}_S^2)} \frac{1}{N_M} \sum_i \mu_i \| f_S(\mathbf{x}_{Si}^1) - f_S(\mathbf{x}_{Si}^2) \|_2 \\
& \ \ \ \ + 2 Z_\phi \mathbb{E}_{\mu, (\mathbf{x}_S^2, \mathbf{x}_M^2)} \frac{1}{N_M} \sum_i \mu_i \| f_S(\mathbf{x}_{Si}^2) - \phi_M(\mathbf{x}_{Mi}^2) \|_2. \\
\end{split}
\end{equation}
\end{thm}
\begin{IEEEproof}
According to the result of Bartlett and Mendelson \cite{PL-Bartlett-and-S-Mendelson-JMLR-2002} and Talagrand contraction Lemma \cite{M-Ledoux-and-M-Talagrand-Springer-2013}, we immediately have the first inequality. The second inequality holds because of the triangle inequality.
\end{IEEEproof}

Note that if the terms $\|\phi_M(\mathbf{x}_{Mi}^1) - f_S(\mathbf{x}_{Si}^1)\|_2$ and $\|f_S(\mathbf{x}_{Si}^2) - \phi_M(\mathbf{x}_{Mi}^2)\|_2$ vanishes, the upper bound of the generalization error for the metric learning algorithm in the target domain will have the same hypothesis complexity as that $\mathbb{E}_{\mu, (\mathbf{x}_S^1, \mathbf{x}_S^2)} \mathop{\mathrm{sup}}_{f_S} \frac{1}{N_S} \sum_i \mu_i \|f_S(\mathbf{x}_{Si}^1) - f_S(\mathbf{x}_{Si}^2)\|_2$ for the metric learning algorithm in the source domain, which is believed to be small because the features are ``expensive'' or there are a large amount of weakly-supervised data in the source domain. This motivates us to penalize $\|\phi_M(\mathbf{x}_{Mi}^1) - f_S(\mathbf{x}_{Si}^1)\|_2$ and $\|f_S(\mathbf{x}_{Si}^2) - \phi_M(\mathbf{x}_{Mi}^2)\|_2$ to be small so that the metric learned in the target domain will inherent the generalization ability of the metric learned in the source domain with ``expensive'' features or a large amount of labeled data.

Note that for any vector $\mathbf{x}$, we have $\| \mathbf{x} \|_2 \leq \| \mathbf{x} \|_1$. To ensure $\|\phi_M(\mathbf{x}_{Mi}^1) - f_S(\mathbf{x}_{Si}^1)\|_2$ and $\|f_S(\mathbf{x}_{Si}^2) - \phi_M(\mathbf{x}_{Mi}^2)\|_2$ to be small, we penalize their upper bounds and introduce the penalty $\frac{\gamma}{N^U} |\Phi_M - F_S|$ in the proposed model, where $\gamma$ is a hyper-parameter. The effectiveness of the proposed model is also verified by empirical experiments. In the following, we first assume $\phi_M = U_M \in \mathbb{R}^{d_M \times r}$ is a linear transformation, and then extend it to the nonlinear case.

\subsection{Linear formulation and optimization}
When we choose $\phi_M = U_M$ as a linear transformation, the problem (\ref{eq:HTDML_KFT_Specific}) can be reformulated as
\begin{equation}
\label{eq:HTDML_KFT_Lin}
\begin{split}
\mathop{\arg \min}_{U_M} \epsilon(U_M)
= & \frac{1}{N_M} \sum_i g\left( y_{Mi} [1 - \| U_M^T(\mathbf{x}_{Mi}^1 - \mathbf{x}_{Mi}^2) \|_2^2] \right) \\
& + \frac{\gamma}{N^U} |U_M^T X_M^U - F_S| \\
& + \frac{\gamma_I}{(N^U)^2} \sum_{i,j} w_{ij} \| U_M^T(\mathbf{x}_{Mi}^U - \mathbf{x}_{Mj}^U) \|_2^2, \\
\mathrm{s.t.} & \ U_M \succeq 0,
\end{split}
\end{equation}
where $X_M^U = [ \mathbf{x}_{M1}^U, \mathbf{x}_{M2}^U, \ldots, \mathbf{x}_{MN^U}^U ] \in \mathbb{R}^{d_M \times N^U}$ is the data matrix of the unlabeled samples in the target domain. The constraint $U_M \succeq 0$ means that each element of $U_M$ is non-negative. This constraint not only narrows the hypothesis space for $U_M$, but also makes the results easy to inspect and interpret.

For notation simplicity, we set $\mathbf{\delta}_{Mi} = \mathbf{x}_{Mi}^1 - \mathbf{x}_{Mi}^2$, so that $\| U_M^¦³ (\mathbf{x}_{Mi}^1 - \mathbf{x}_{Mi}^2) \|_2^2 = \delta_{Mi}^T U_M U_M^T \delta_{Mi}$. Then the optimization problem becomes
\begin{equation}
\label{eq:HTDML_KFT_wrt_U}
\begin{split}
\mathop{\arg \min}_{U_M} \epsilon(U_M) = E (U_M) + \Omega (U_M),\ \mathrm{s.t.}\ U_M \succeq 0,
\end{split}
\end{equation}
where $E(U_M) = \frac{1}{N_M} \sum_{i=1}^{N_M} g(y_{Mi} [1 - \delta_{Mi}^T U_M U_M^T \delta_{Mi}])$ and
\begin{equation}
\notag
\begin{split}
\Omega(U_M) = & \frac{\gamma}{N^U} |U_M^T X_M^U - F_S| \\
& + \frac{\gamma_I}{(N^U)^2} \mathrm{tr}(U_M^T X_M^U L_M (X_M^U)^T U_M).
\end{split}
\end{equation}
The matrix $L_M$ is the graph Laplacian as defined in \cite{M-Belkin-et-al-JMLR-2006}. We propose to solve the problem (\ref{eq:HTDML_KFT_wrt_U}) efficiently by utilizing the projected gradient method (PGM) presented in \cite{CJ-Lin-NCn-2007}. Because both $E(U_M)$ and the first term in $\Omega(U_M)$ are non-differentiable, we first smooth it according to the Nesterov's smoothing technique \cite{Y-Nesterov-MP-2005}. According to \cite{Y-Nesterov-MP-2005}, the smoothed version of the hinge loss $g(U_M; \delta_{Mi}, y_{Mi}) = \mathrm{\max}\{ 0, -y_{Mi} (1 - \delta_{Mi}^T U_M U_M^T \delta_{Mi}) \}$ can be given by
\begin{equation}
\label{eq:Hinge_Loss_Smooth}
\begin{split}
& g^\sigma (U_M; \delta_{Mi}, y_{Mi}) \\
& = \mathop{\max}_{\nu \in \mathcal{Q}} \nu_i \left( - y_{Mi} (1 - \delta_{Mi}^T U_M U_M^T \delta_{Mi}) \right) - \frac{\sigma}{2} \| \delta_{Mi} \|_\infty \nu_i^2,
\end{split}
\end{equation}
where $\mathcal{Q} = \{ \nu: 0 \leq \nu_i \leq 1, \nu \in \mathbb{R}^N \}$ and $\sigma$ is the smooth hyper-parameter, which should be neither too large or too small. A larger $\sigma$ corresponding to a smoother approximation but larger approximation error. When $\sigma$ is small, the convergence rate is slow and thus the time complexity is high. We set it as $0.5$ according to the empirical study in \cite{TY-Zhou-et-al-ICDM-2010}. By setting the gradient of the objective function in (\ref{eq:Hinge_Loss_Smooth}) to become zero and then projecting $\nu_i$ on $\mathcal{Q}$, we obtain the following solution,
\begin{equation}
\label{eq:Hinge_Smooth_Solution}
\nu_i = \mathrm{median} \left\{ \frac{- y_{Mi} (1 - \delta_{Mi}^T U_M U_M^T \delta_{Mi})}{\sigma \| \delta_{Mi} \|_\infty}, 0, 1 \right\}.
\end{equation}
By substituting the solution (\ref{eq:Hinge_Smooth_Solution}) back into (\ref{eq:Hinge_Loss_Smooth}), we have the piece-wise approximation of $g$, i.e.,
\begin{equation}
\label{eq:Hinge_Loss_Approximation}
\begin{split}
& g^\sigma = \\
& \left\{
\begin{array}{cc}
0, & \begin{split} y_{Mi} (1 - \Delta_{Mi}) > 0 \end{split}; \\
- y_{Mi} (1 - \Delta_{Mi}) - \frac{\sigma}{2} \| \delta_{Mi} \|_\infty, & \begin{split} y_{Mi} (1 - \Delta_{Mi}) \\ < - \sigma \| \delta_{Mi} \|_\infty \end{split}; \\
\frac{\left( y_{Mi} (1 - \Delta_{Mi}) \right)^2}{2 \sigma \| \delta_{Mi} \|_\infty}, & \mathrm{otherwise},
\end{array}
\right.
\end{split}
\end{equation}
where $\Delta_{Mi} = \delta_{Mi}^T U_M U_M^T \delta_{Mi}$. To utilize PGM for optimization, we have to compute the gradient of the smoothed hinge loss to determine the direction of the descent. We summarize the results in the following theorem.
\begin{thm}
\label{thm:Hinge_Smooth_Gradient}
The sum of gradient of the smoothed hinge loss $g^\sigma$ w.r.t. $U_M$ over all samples is
\begin{equation}
\label{eq:Hinge_Smooth_Gradient}
\frac{\partial g^\sigma(U_M)}{\partial U_M} = \sum_i \left( 2 y_{Mi} \nu_i (\delta_{Mi} \delta_{Mi}^T) U_M \right).
\end{equation}
Here, $\nu_i$ is related to $U_M$.
\end{thm}
We leave the proof in the supplementary material.

Similarly, for the sum of $l_1$-norm $h(U_M) = |U_M^T X_M^U - F_S| = \sum_{c=1}^r \sum_{n=1}^{N^U} h(\mathbf{u}_{Mc}^T \mathbf{x}_{Mn}^U - f_{S,cn})$, where $h(z) = |z|$, $\mathbf{u}_{Mc}$ is the $c$-th column of $U_M$ and $f_{S,cn}$ is the $(c,n)$-th element of $F_S$, we have the following smoothed version:
\begin{equation}
\label{eq:L1_Norm_Smooth}
h^\sigma(\mathbf{u}_{Mc}^T \mathbf{x}_{Mn}^U - f_{S,cn}) = \mathop{\max}_{Q \in \mathcal{Q}} \langle \mathbf{u}_{Mc}^T \mathbf{x}_{Mn}^U - f_{S,cn}, q_{cn} \rangle - \frac{\sigma}{2} q_{cn}^2,
\end{equation}
where $\mathcal{Q} = \{ Q: -1 \leq q_{cn} \leq 1, Q \in \mathbb{R}^{r \times {N^U}} \}$, and $\sigma$ is the smooth hyper-parameter. By setting the objective function of (\ref{eq:L1_Norm_Smooth}) as zero and then projecting $q_{cn}$ on $\mathcal{Q}$, we obtain the following solution:
\begin{equation}
\label{eq:L1_Smooth_Solution}
q_{cn} = \mathrm{median} \left\{ \frac{\mathbf{u}_{Mc}^T \mathbf{x}_{Mn}^U - f_{S,cn}}{\sigma}, -1, 1 \right\}.
\end{equation}
By substituting the solution (\ref{eq:L1_Smooth_Solution}) back into (\ref{eq:L1_Norm_Smooth}), we have the following piece-wise approximation of $h$, i.e.,
\begin{equation}
\label{eq:L1_Norm_Approximation}
\begin{split}
& h^\sigma = \\
& \left\{
\begin{array}{cc}
- (\mathbf{u}_{Mc}^T \mathbf{x}_{Mn}^U - f_{S,cn}) - \frac{\sigma}{2}, & \mathbf{u}_{Mc}^T \mathbf{x}_{Mn}^U - f_{S,cn} < - \sigma; \\
(\mathbf{u}_{Mc}^T \mathbf{x}_{Mn}^U - f_{S,cn}) - \frac{\sigma}{2}, & \mathbf{u}_{Mc}^T \mathbf{x}_{Mn}^U - f_{S,cn} > \sigma; \\
\frac{(\mathbf{u}_{Mc}^T \mathbf{x}_{Mn}^U - f_{S,cn})^2}{2 \sigma}, & \mathrm{otherwise}.
\end{array}
\right.
\end{split}
\end{equation}
The gradient of smoothed $h(U_M) = |U_M^T X_M^U - F_S|$ is given by the following theorem:
\begin{thm}
\label{thm:L1_Smooth_Gradient}
The gradient of the smoothed $|U_M^T X_M^U - F_S|$, i.e., $h^\sigma$ w.r.t. $U_M$ is
\begin{equation}
\label{eq:L1_Smooth_Gradient}
\frac{\partial h^\sigma(U_M)}{\partial U_M} = \frac{\partial |U_M^T X_M^U - F_S|}{\partial U_M} = X_M^U Q^T,
\end{equation}
where $Q$ is a matrix related to $U_M$ with the entry $q_{cn}$ given by (\ref{eq:L1_Smooth_Solution}).
\end{thm}
We leave the proof in the supplementary material.

Therefore, the gradient of the smoothed $\epsilon(U_M)$ is
\begin{equation}
\label{eq:All_Smooth_Gradient}
\begin{split}
\frac{\partial \epsilon^\sigma (U_M)}{\partial U_M} = & \frac{1}{N_M} \sum_{i=1}^{N_M} \left( 2 y_{Mi} \nu_i (\delta_{Mi} \delta_{Mi}^T) U_M \right) \\
& + \frac{\gamma}{N^U} X_M^U Q^T + \frac{\gamma_I}{(N^U)^2} (2 X_M^U L_M (X_M^U)^T U_M).
\end{split}
\end{equation}
Finally, based on the obtained gradient, we apply the improved PGM presented in \cite{CJ-Lin-NCn-2007} to minimize the smoothed primal $\epsilon^\sigma (U_M)$, i.e.,
\begin{equation}
\label{eq:PGM_Update_Rule}
U_M^{t+1} = \pi [U_M^t - \mu_t \nabla \epsilon^\sigma (U_M^t)],
\end{equation}
where the operator $\pi [x]$ projects all the negative entries of $x$ to zero, and $\mu_t$ is the step size that must satisfy the following condition:
\begin{equation}
\label{eq:PGM_Step_Size}
\epsilon^\sigma(U_M^{t+1}) - \epsilon^\sigma(U_M^t) \leq \rho \nabla \epsilon^\sigma(U_M^t)^T (U_M^{t+1} - U_M^t),
\end{equation}
where the hyper-parameter $\rho$ is chosen to be $0.01$ following \cite{CJ-Lin-NCn-2007}. The step size can be determined using the Algorithm 4 in \cite{CJ-Lin-NCn-2007}, where the convergence of the algorithm is guaranteed. The stopping criterion we utilized here is $|\epsilon^\sigma (U_M^{t+1}) - \epsilon^\sigma (U_M^t)| / |\epsilon^\sigma (U_M^{t+1}) - \epsilon^\sigma (U_M^0)| < \varepsilon)$, where the initialization $U_M^0$ is the set as a random matrix.

\subsection{Nonlinear extension}
When we allow the mapping $\phi_M$ to be nonlinear, the problem (\ref{eq:HTDML_KFT_Specific}) can be rewritten as
\begin{equation}
\label{eq:HTDML_KFT_Non}
\begin{split}
& \mathop{\arg \min}_{\phi_M} \epsilon(\phi_M) \\
& = \frac{1}{N_M} \sum_{i=1}^{N_M} g\left( y_{Mi} [1 - \| \phi_M(\mathbf{x}_{Mi}^1) - \phi_M(\mathbf{x}_{Mi}^2) \|_2^2] \right) \\
& \ \ \ \ + \frac{\gamma}{N^U} \sum_{n=1}^{N^U} |\phi_M (\mathbf{x}_{Mn}^U) - \mathbf{f}_{Sn}| \\
& \ \ \ \ + \frac{\gamma_I}{(N^U)^2} \sum_{i,j=1}^{N^U} w_{ij} \| \phi_M(\mathbf{x}_{Mi}^U) - \phi_M(\mathbf{x}_{Mj}^U) \|_2^2,
\end{split}
\end{equation}
where $\mathbf{f}_{Sn}$ is the $n$-th column of the matrix $F_S$. To find an appropriate nonlinear form for $\phi_M$, we assume it is a gradient boosting function given by $\phi_M = \phi_M^0 + \alpha \sum_{t=1}^T \hbar_{Mt}$. Here, $\phi_M^0$ is an initialization, and $\hbar_{Mt}$ is a regression tree together with a learning rate $\alpha$ \cite{D-Kedem-et-al-NIPS-2012}. The solution can be obtained by iteratively adding regression trees $\hbar_{Mt}$ to minimize the objective $\epsilon(\phi_M)$ in a greedy way \cite{JH-Friedman-AoS-2001}.

In the following, we summarize the procedure of finding the (approximately) optimal tree in each iteration. In iteration $t$, the (approximately) optimal tree $\hbar_{Mt}^\ast$ is found by selecting a tree from the set of all regression trees $\mathcal{T}^p$ to approximate the negative gradient of $\epsilon(\phi_M^{t-1})$ w.r.t. $\phi_M^{t-1}$, which is the mapping learned at the previous iteration. Here, $p$ is the depth of the trees. Similar to the linear formulation, we smooth the non-differentiable terms to calculate the gradients. The smoothed hinge loss is given by (\ref{eq:Hinge_Loss_Approximation}), where the term $\delta_{Mi}^T U_M U_M^T \delta_{Mi}$ is replaced with the general form $\| \phi_M(\mathbf{x}_{Mi}^1) - \phi_M(\mathbf{x}_{Mi}^2) \|_2^2$. The smoothed $l_1$-norm, $|\phi_{Mc} (\mathbf{x}_{Mn}^U) - f_{S,cn}|$ is given by (\ref{eq:L1_Norm_Approximation}), where the term $\mathbf{u}_{Mc}^T \mathbf{x}_{Mn}^U$ is replaced with $\phi_{Mc} (\mathbf{x}_{Mn}^U)$, which is $c$-th entry of the vector $\phi_M (\mathbf{x}_{Mn}^U)$, and $f_{S,cn}$ is the $c$-th entry of $\mathbf{f}_{Sn}$. Then the tree $\hbar_{Mt}^\ast(\cdot)$ is learned by approximating it with the negative gradient $neg_t(\cdot)$ over each training sample, i.e.,
\begin{equation}
\label{eq:Tree_Approximation}
\begin{split}
\hbar_{Mt}^\ast(\cdot) = & \mathop{\arg \min}_{\hbar \in \mathcal{T}^p} \sum_{i=1}^{N_M} \left( \hbar(\mathbf{x}_{Mi}) - neg_t(\mathbf{x}_{Mi}) \right)^2 \\
& + \sum_{n=1}^{N^U} \left( \hbar(\mathbf{x}_{Mn}^U) - neg_t(\mathbf{x}_{Mn}^U) \right)^2,
\end{split}
\end{equation}
where $neg_t(\mathbf{x}_{Mi}) = - \frac{\partial \epsilon(\phi_M^{t-1})}{\partial \phi_M^{t-1}(\mathbf{x}_{Mi})}$. In particular, for a given triplet $\langle \mathbf{x}_{Mi}^1, \mathbf{x}_{Mi}^2, y_{Mi} \rangle$, the gradients are calculated as
\begin{equation}
\notag
\frac{\partial \epsilon(\phi_M^{t-1})}{\partial \phi_M^{t-1} (\mathbf{x}_{Mi}^1)} = \frac{2}{N_M} y_{Mi} \nu_i \left( \phi(\mathbf{x}_{Mi}^1) - \phi(\mathbf{x}_{Mi}^2) \right)
\end{equation}
and
\begin{equation}
\notag
\frac{\partial \epsilon(\phi_M^{t-1})}{\partial \phi_M^{t-1}(\mathbf{x}_{Mi}^2)} = \frac{2}{N_M} y_{Mi} \nu_i \left( \phi(\mathbf{x}_{Mi}^2) - \phi(\mathbf{x}_{Mi}^1) \right),
\end{equation}
where $\nu_i = \mathrm{median}\{ \frac{ - y_{Mi} (1 - \| \phi(\mathbf{x}_{Mi}^1) - \phi(\mathbf{x}_{Mi}^2) \|_2^2) }{\sigma \| \delta_{Mi} \|_\infty}, 0, 1 \}$. For a given unlabeled sample, the gradient consists of two parts. The first part is
\begin{equation}
\notag
\frac{\partial \epsilon(\phi_M^{t-1})}{\partial \phi_M^{t-1}(\mathbf{x}_{Mn}^U)} = \frac{\gamma}{N^U} \mathbf{q}_n,
\end{equation}
where $\mathbf{q}_n$ is a vector with the entry $q_{cn} = \mathrm{median}\{ \frac{\phi_{Mc} (\mathbf{x}_{Mn}^U) - f_{S,cn}}{\sigma}, -1, 1 \}$. The second part is
\begin{equation}
\notag
\frac{\partial \epsilon(\phi_M^{t-1})}{\partial \phi_M^{t-1}(\mathbf{x}_{Mi}^U)} = \frac{2\gamma_I}{(N^U)^2} \sum_j w_{ij} \left( \phi(\mathbf{x}_{Mi}^U) - \phi(\mathbf{x}_{Mj}^U) \right)
\end{equation}
and
\begin{equation}
\notag
\frac{\partial \epsilon(\phi_M^{t-1})}{\partial \phi_M^{t-1}(\mathbf{x}_{Mj}^U)} = \frac{2\gamma_I}{(N^U)^2} \sum_i w_{ij} \left( \phi(\mathbf{x}_{Mj}^U) - \phi(\mathbf{x}_{Mi}^U) \right).
\end{equation}
The tree is greedily learned by pGBRT \cite{S-Tyree-et-al-WWW-2011}. The problem (\ref{eq:HTDML_KFT_Non}) is non-convex w.r.t. $\phi_M$, so we initialize $\phi_M$ as $\phi_M^0 = U_M^\ast$, which is the optimal transformation learned by our linear formulation. This makes the extension to be a nonlinear refinement of the linear formulation. A complexity analysis of the proposed method can be found in the supplementary material.

\section{Experiments}\label{sec:Experiments}

In this section, we evaluate the effectiveness of the proposed HTDML algorithm on both scene classification and object recognition. Prior to these evaluations, we present our experimental settings.

\subsection{Experimental setup}
The comparison methods are listed as below:
\begin{itemize}
  \item \textbf{EU:} directly computing the Euclidean distance between the normalized representations of different samples in the target domain.
  \item \textbf{LMNN \cite{KQ-Weinberger-et-al-NIPS-2005}:} learning the distance metric for the target domain using the large margin nearest neighbor algorithm presented in \cite{KQ-Weinberger-et-al-NIPS-2005}. The number of attracted target neighbors is chosen from $1$ to $10$.
  \item \textbf{ITML \cite{JV-Davis-et-al-ICML-2007}:} learning the distance metric for the target domain using the information-theoretic metric learning algorithm presented in \cite{JV-Davis-et-al-ICML-2007}. The trade-off hyper-parameter is tuned over the set $\{ 10^i | i = -5, -4, \ldots , 3, 4 \}$.
  \item \textbf{MTDA \cite{Y-Zhang-and-DY-Yeung-AAAI-2011}:} a heterogeneous multi-task learning algorithm by extending linear discriminant analysis to handle multiple heterogeneous domains. The hyper-parameter of intermediate dimensionality is set as a fixed value since the model is not very sensitive to it according to \cite{Y-Zhang-and-DY-Yeung-AAAI-2011}.
  \item \textbf{DAMA \cite{C-Wang-and-S-Mahadevan-IJCAI-2011}:} a heterogeneous domain adaptation algorithm by aligning the manifolds of different domains using the class labels, and simultaneously preserving the topology of each domain using large amounts of unlabeled data. The hyper-parameter is determined according to the strategy presented in \cite{C-Wang-and-S-Mahadevan-IJCAI-2011}.
  \item \textbf{DT \cite{GJ-Qi-et-al-SDM-2012}:} a heterogeneous distance function transfer algorithm by leveraging large amounts of corresponding data between the source and target domain. The similarity of two target domain samples is determined according to the similarities of their corresponding source domain samples. The candidate set for the two balancing hyper-parameters are both $\{ 10^i | i = -5, -4, \ldots, 4 \}$.
  \item \textbf{MI \cite{DX-Dai-et-al-CVPR-2015}:} a recently proposed metric imitation algorithm that utilizes the ``expensive'' features to guide the metric learning of some ``cheap'' features. The knowledge transfer is performed by manifold structure approximation between the source and target domains. We adopt LapEigen \cite{M-Belkin-and-P-Niyogi-NIPS-2001} to characterize the local manifold structure in the source domain. The hyper-parameters are determined according to \cite{DX-Dai-et-al-CVPR-2015}.
  \item \textbf{HTDML:} the proposed heterogeneous transfer distance metric learning algorithm. In the linear formulation, LMNN \cite{KQ-Weinberger-et-al-NIPS-2005} is adopted to find the fundamental elements in the source domain. In the nonlinear extension, we employ GB-LMNN \cite{D-Kedem-et-al-NIPS-2012} to learn the source knowledge fragments. Because GBRT is adopted to learn the mapping in the target domain, we call the proposed nonlinear extension \textbf{GB-HTDML}. Both the hyper-parameters $\gamma$ and $\gamma_I$ are optimized over the set $\{ 10^i | i = -3, -4, \ldots, 5, 6 \}$.
\end{itemize}

The single domain distance metric learning (DML) algorithms (LMNN and ITML) only utilize the given limited label information in each domain, and do not make use of any additional information from other domains. The heterogeneous transfer learning (HTL) approaches, MTDA and DAMA, mainly utilize the label information in both the source and target domain to build a connection between them. DAMA also leverages large amounts of unlabeled data to preserve the topology in each domain. They do not aim to learn distance metric, so we derive the metric as $A = U U^T$ after learning the transformation matrix $U \in \mathbb{R}^{d \times r}$ for the target domain. DT and MI are also HTL methods, but they focus on metric learning and perform knowledge transfer by utilizing the unlabeled correspondence information between the source and target domain. The proposed HTDML aims to learn an improve distance metric in the target domain by making use of both the additional label information from source domain, as well as the unlabeled correspondence. The chosen of an optimal dimensionality $r$ of the mapped subspace is still an open problem, and we do not study it in this paper. To this end, for MTDA, DAMA, and the proposed HTDML, we perform performance comparisons on a set of varied $r$. In all the following experiments, each feature space is regarded as a domain. Hyper-parameter determination is still an open issue in heterogenous transfer learning \cite{JT-Zhou-et-al-AISTATS-2014} due to the limited labeled samples in the target domain. Consequently, if unspecified, the hyper-parameters are tuned in the range mentioned above and the best results of different compared methods are reported.

\subsection{Scene categorization}
The dataset used in scene categorization is the Scene-15 \cite{S-Lazebnik-et-al-CVPR-2006}, which contains $4585$ images belonging to $15$ natural scene categories. We randomly split the image set into a training and test set of equal size. We choose the ``expensive'' CNN feature \cite{K-Chatfield-et-al-arXiv-2014} as the source domain, and the ``cheap'' GIST feature \cite{A-Oliva-and-A-Torralba-IJCV-2001} as the target domain. The used features are provided by \cite{DX-Dai-et-al-CVPR-2015}, where the feature dimensions of CNN and GIST are $4096$ and $20$ respectively. The task in the target domain is to perform multi-class classification, where the $k$-nearest neighbor classifier is adopted. We select $10$ labeled instances for each category in both the source and target domain to see how much the ``expensive'' feature can help the metric learning of ``cheap'' feature \cite{DX-Dai-et-al-CVPR-2015}. Both the classification accuracy and macroF1 \cite{M-Sokolova-and-G-Lapalme-IPM-2009} score are utilized as evaluation criteria. The label information in terms of pairwise similarity constraints are obtained according to whether two labeled training samples belong to the same class or not. The remained training data that have representations in both domains are used as unlabeled data. Ten random choices of the labeled instances or sample pairs are used, and the mean values with standard deviations are reported. In the following, we first select $2000$ unlabeled cross-domain correspondences, and then investigate the performance of varying number of correspondences.

\begin{figure}
\centering
\subfloat{\includegraphics[width=0.485\columnwidth]{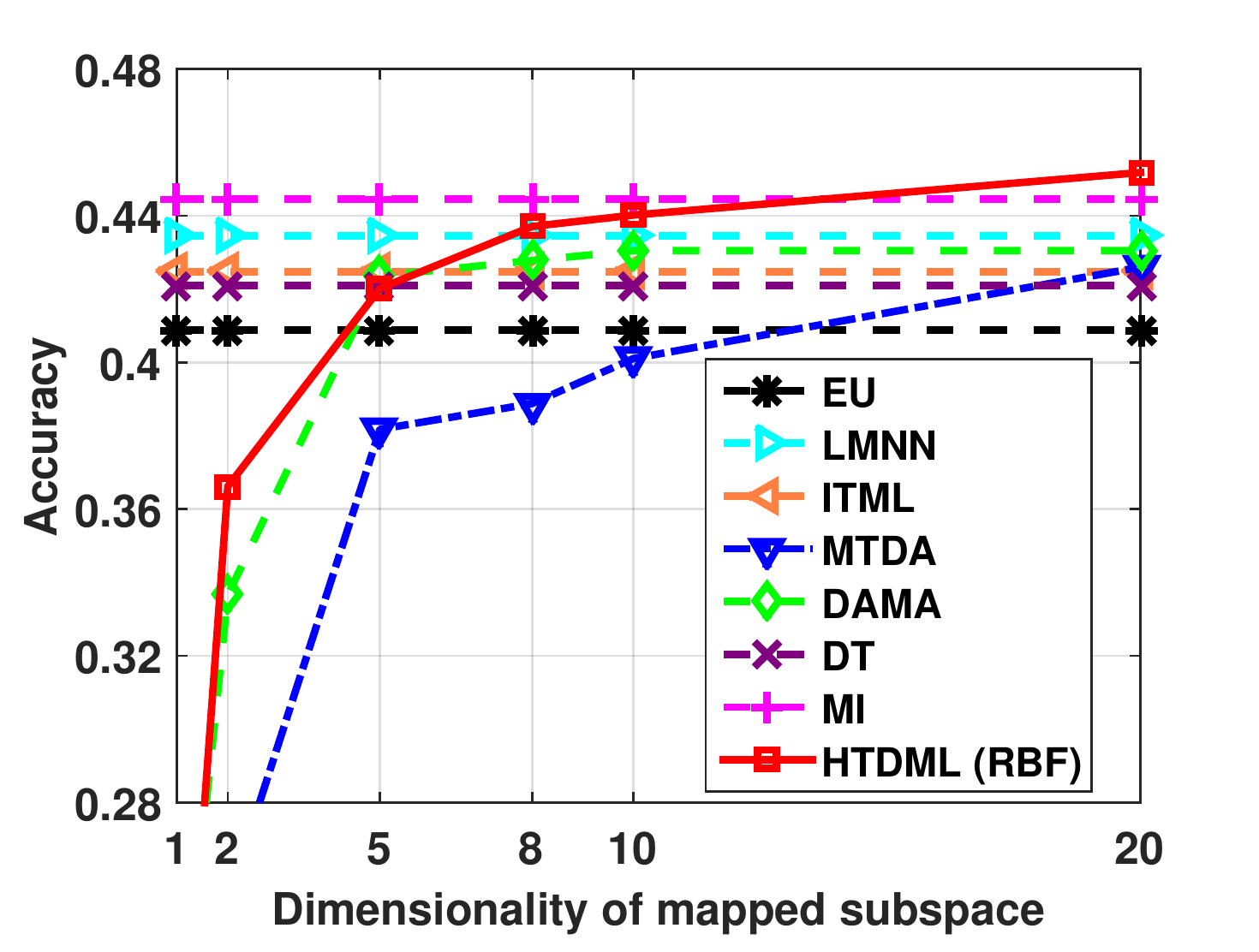}
}
\hfil
\subfloat{\includegraphics[width=0.485\columnwidth]{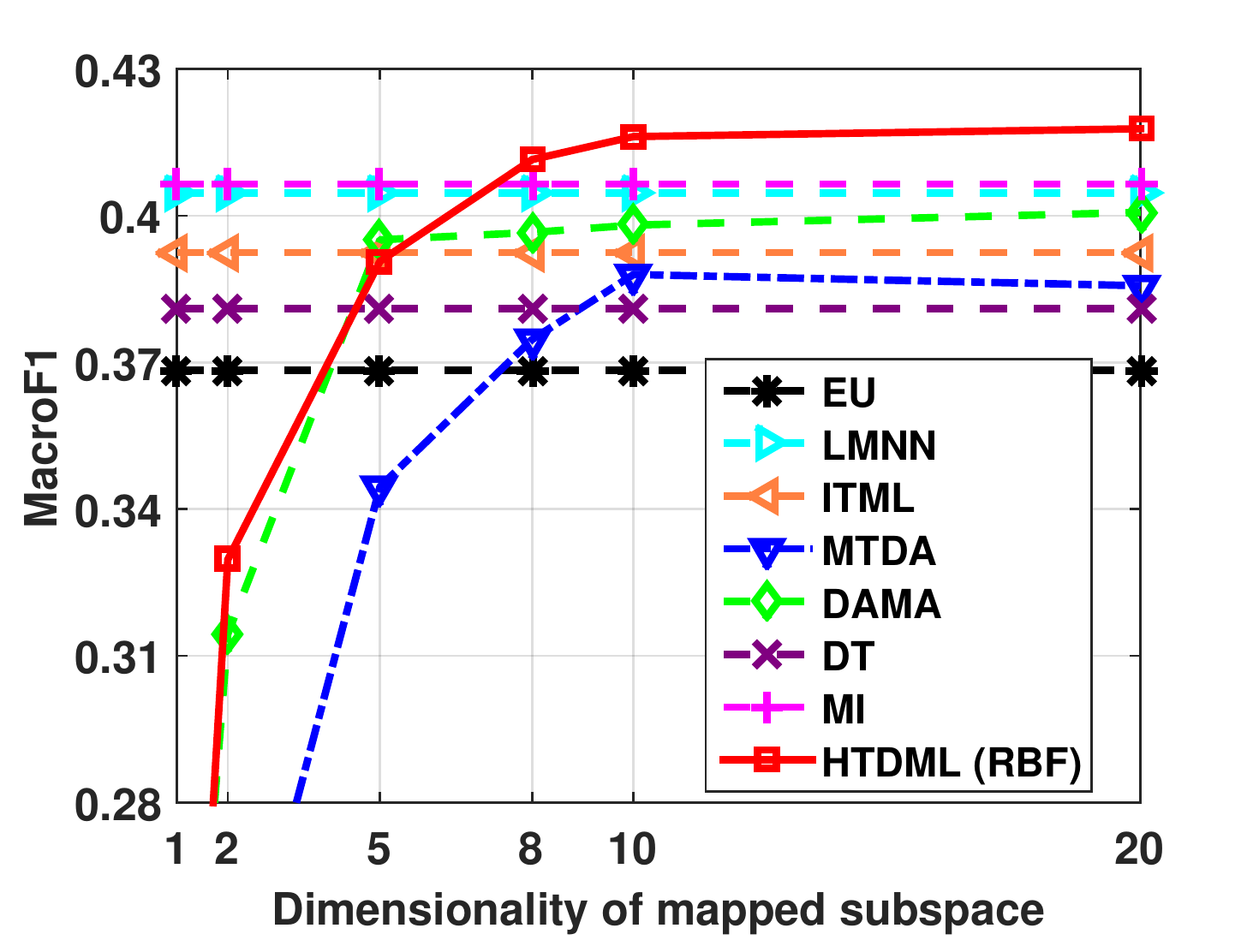}
}
\caption{Classification accuracies and macroF1 scores vs. dimensionality of the mapped subspace on the Scene-15 dataset.}
\label{fig:Acc_MacF1_vs_Dim_SCE_GIST_Lin}
\end{figure}

\begin{table}[!t]
\renewcommand{\arraystretch}{1.3}
\caption{Classification accuracies and macroF1 scores of the compared methods at their best dimensionalities of mapped subspace on the Scene-15 dataset.}
\label{tab:Acc_MacF1_Bst_SCE_GIST_Lin}
\centering
\begin{tabular}{c||c|c}
\hline
Methods & Accuracy & MacroF1 \\
\hline
EU & 0.409$\pm$0.017 & 0.368$\pm$0.016 \\
\hline
LMNN & 0.435$\pm$0.020 & 0.405$\pm$0.022 \\
ITML & 0.425$\pm$0.021 & 0.393$\pm$0.022 \\
\hline
MTDA & 0.426$\pm$0.074 & 0.388$\pm$0.014 \\
DAMA & 0.431$\pm$0.015 & 0.401$\pm$0.012 \\
\hline
DT & 0.421$\pm$0.014 & 0.381$\pm$0.014 \\
MI & 0.445$\pm$0.023 & 0.407$\pm$0.020 \\
\hline
HTDML (RBF) & \textbf{0.452$\pm$0.013} & \textbf{0.418$\pm$0.014} \\
\hline
\end{tabular}
\end{table}

\subsubsection{Evaluation of the linear and nonlinear formulation}
In our linear formulation, we choose the RBF kernel $\kappa(\mathbf{x}_i, \mathbf{x}_j) = \mathrm{exp}(- \| \mathbf{x}_i - \mathbf{x}_j \|^2 / (2 \omega^2))$ to construct the knowledge fragments in the source domain. The hyper-parameter $\omega$ of the RBF kernel is set as the mean distance between all neighbors, i.e., $\omega = \frac{1}{(N^U)^2} \sum_i \sum_j \| \mathbf{x}_i - \mathbf{x}_j \|^2$. The classification accuracies and macroF1 scores in relation to the number $r$ are shown in Fig. \ref{fig:Acc_MacF1_vs_Dim_SCE_GIST_Lin}. The performance of different methods at their best dimensionalities are summarized in Table \ref{tab:Acc_MacF1_Bst_SCE_GIST_Lin}. From these results, we observe that: 1) although the label information is limited in the target domain, the single domain DML algorithms (LMNN and ITML) can still improve the performance; 2) the supervised heterogeneous transfer learning (HTL) approaches (MTDA and DAMA) are superior to the EU baseline by utilizing the label information from both domains. DAMA is better than MTDA since the former makes use of additional unlabeled data to preserve the topology in each domain. But DAMA is only comparable to the single domain DML algorithms at its best dimensionality, and MTDA is even worse than LMNN. This may be because the structures of the visual feature distributions are usually nonlinear. By using only linear transformations, it is hard for the HTL approaches to build a connection between the source and target domain, and thus fail to transfer the information; 3) DT is designed for metric transfer, but it is even worse than the single domain DML algorithms on MacroF1 score, since it does not take advantage of any label information and also suffer from the limitation of linearity; 4) MI is better than all the other methods except the proposed HTDML. This may be because the nonlinear structure of the data distribution is explored in the source domain by the learned manifold; 5) the proposed HTDML outperforms MI since we effectively utilize both the similar/dissimilar constraints and correspondence information.

\begin{figure}
\centering
\subfloat{\includegraphics[width=0.485\columnwidth]{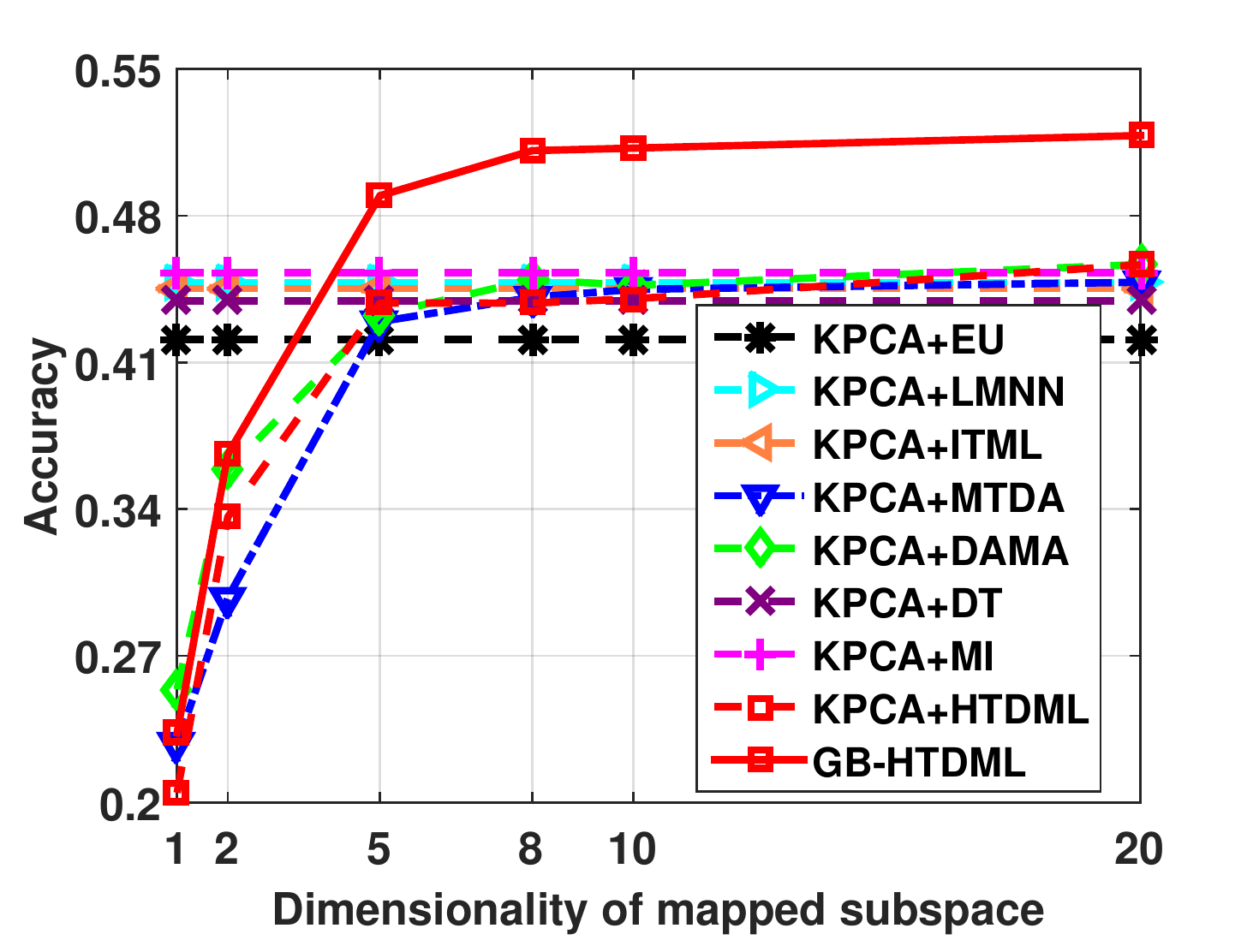}
}
\hfil
\subfloat{\includegraphics[width=0.485\columnwidth]{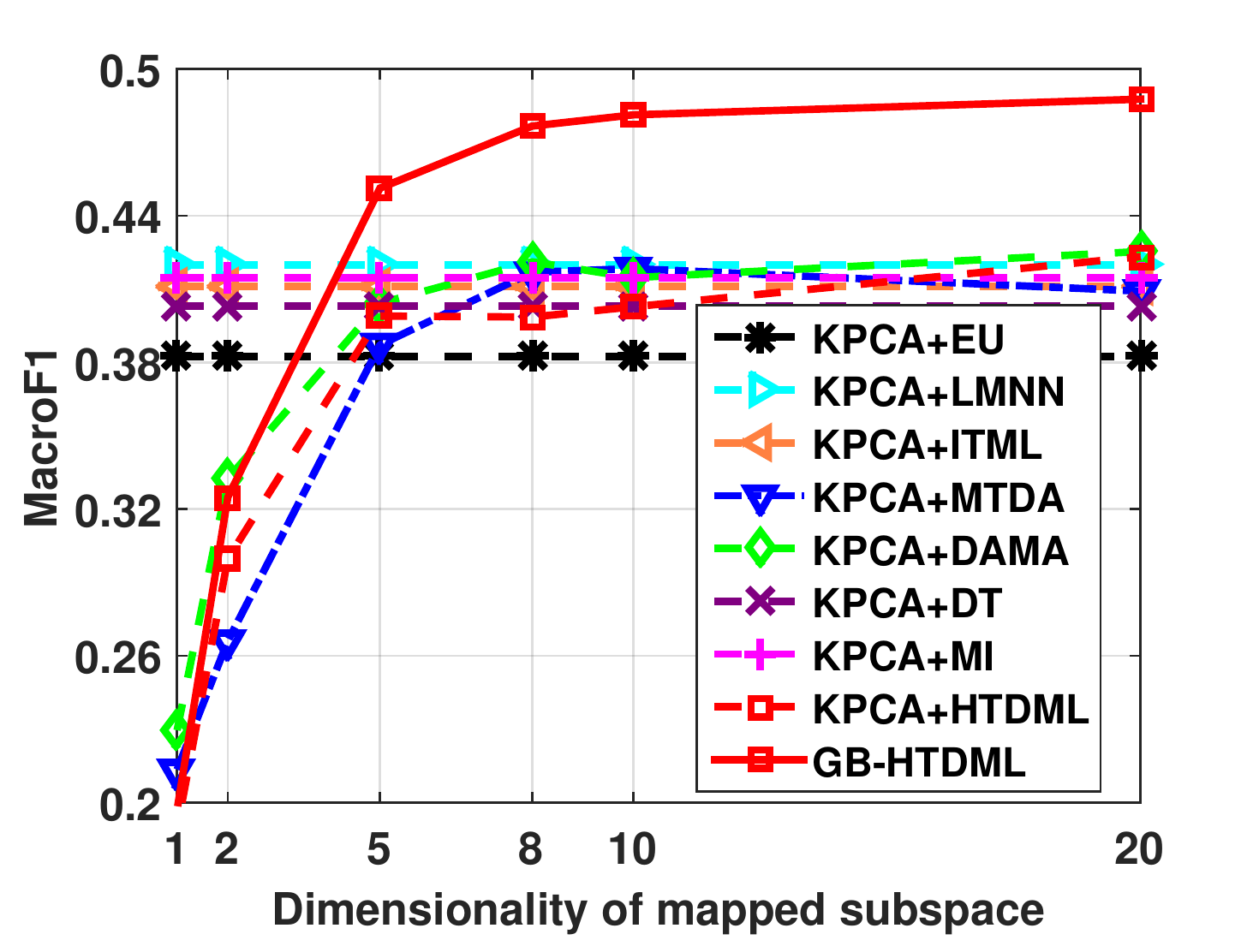}
}
\caption{Classification accuracies and macroF1 scores of the nonlinear methods vs. dimensionality of the mapped subspace on the Scene-15 dataset.}
\label{fig:Acc_MacF1_vs_Dim_SCE_GIST_Non}
\end{figure}

\begin{table}[!t]
\renewcommand{\arraystretch}{1.3}
\caption{Classification accuracies and macroF1 scores of the compared nonlinear methods at their best dimensionalities of mapped subspace on the Scene-15 dataset.}
\label{tab:Acc_MacF1_Bst_SCE_GIST_Non}
\centering
\begin{tabular}{c||c|c}
\hline
Methods & Accuracy & MacroF1 \\
\hline
KPCA+EU & 0.421$\pm$0.022 & 0.383$\pm$0.021 \\
\hline
KPCA+LMNN & 0.448$\pm$0.022 & 0.420$\pm$0.022 \\
KPCA+ITML & 0.445$\pm$0.022 & 0.411$\pm$0.022 \\
\hline
KPCA+MTDA & 0.448$\pm$0.006 & 0.418$\pm$0.004 \\
KPCA+DAMA & 0.457$\pm$0.015 & 0.425$\pm$0.018 \\
\hline
KPCA+DT & 0.439$\pm$0.014 & 0.403$\pm$0.014 \\
KPCA+MI & 0.453$\pm$0.014 & 0.415$\pm$0.017 \\
\hline
KPCA+HTDML & 0.457$\pm$0.013 & 0.423$\pm$0.013 \\
GB-HTDML & \textbf{0.518$\pm$0.012} & \textbf{0.488$\pm$0.017} \\
\hline
\end{tabular}
\end{table}

To evaluate the nonlinear formulation, for all the compared methods except the proposed GB-HTDML, we preprocess the features by kernel PCA (KPCA) to take the nonlinear structure of the data distribution into consideration. In the proposed linear HTDML with the KPCA preprocess, we adopt the linear kernel in the source domain. The result dimensions are $2000$ and $20$ for CNN and GIST respectively. We show the classification results in Fig. \ref{fig:Acc_MacF1_vs_Dim_SCE_GIST_Non}, and summarize the performance of different methods at their best dimensionalities in Table \ref{tab:Acc_MacF1_Bst_SCE_GIST_Non}. We can see from the results that: 1) the performance of all methods are better than the results without KPCA preprocess. This indicates that the distribution structures of the utilized visual features are indeed nonlinear, and KPCA is able to exploit such nonlinearity to some extent; 2) the HTL approach DAMA is slightly better than the single domain DML algorithms. This is because KPCA helps to build the domain connection by exploiting the nonlinearity, and some source information is transferred to help learning the transformation in the target domain; 3) DT and MTDA are only comparable to LMNN and ITML. The performance of MI and the proposed HTDML with the KPCA preprocess is only a bit higher than LMNN. This indicates that such a simple preprocess can only bring limited benefits to the transfer approaches, and is not always helpful; 4) the proposed GB-HTDML outperforms all other approaches significantly. This demonstrates that the nonlinearity in the data is successfully captured by the developed nonlinear algorithm. In particular, we obtain significant $13.5\%$ and $14.6\%$ relative improvements over the competitive DAMA in terms of accuracy and macroF1 respectively.

A computational cost comparison of different methods can be found in the supplementary material. Due to the nonlinear structures of the visual feature distributions, we only investigate or evaluate the nonlinear formulation in the following experiments.

\subsubsection{A self-comparison analysis}
\begin{figure}
\centering
\subfloat{\includegraphics[width=0.485\columnwidth]{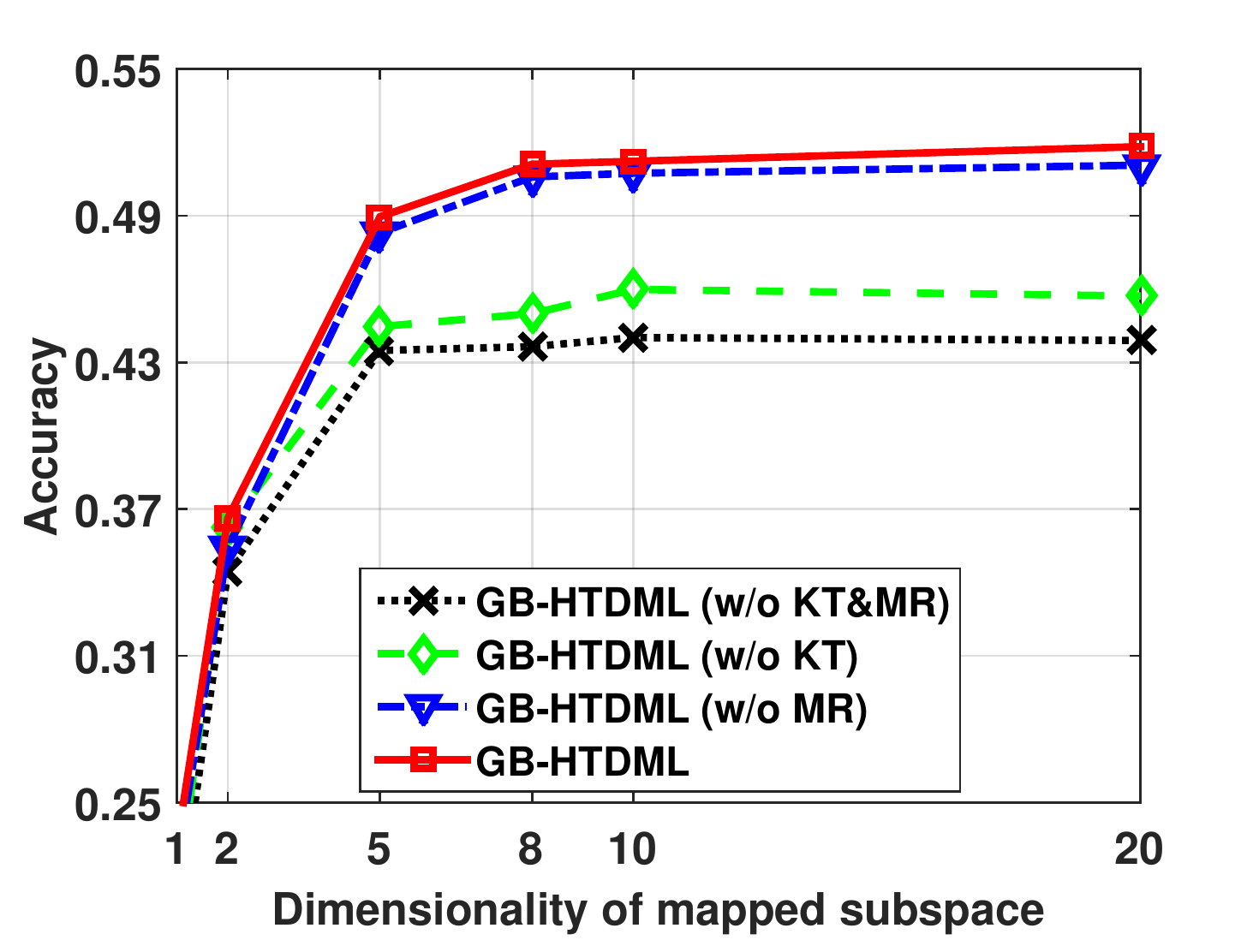}
}
\hfil
\subfloat{\includegraphics[width=0.485\columnwidth]{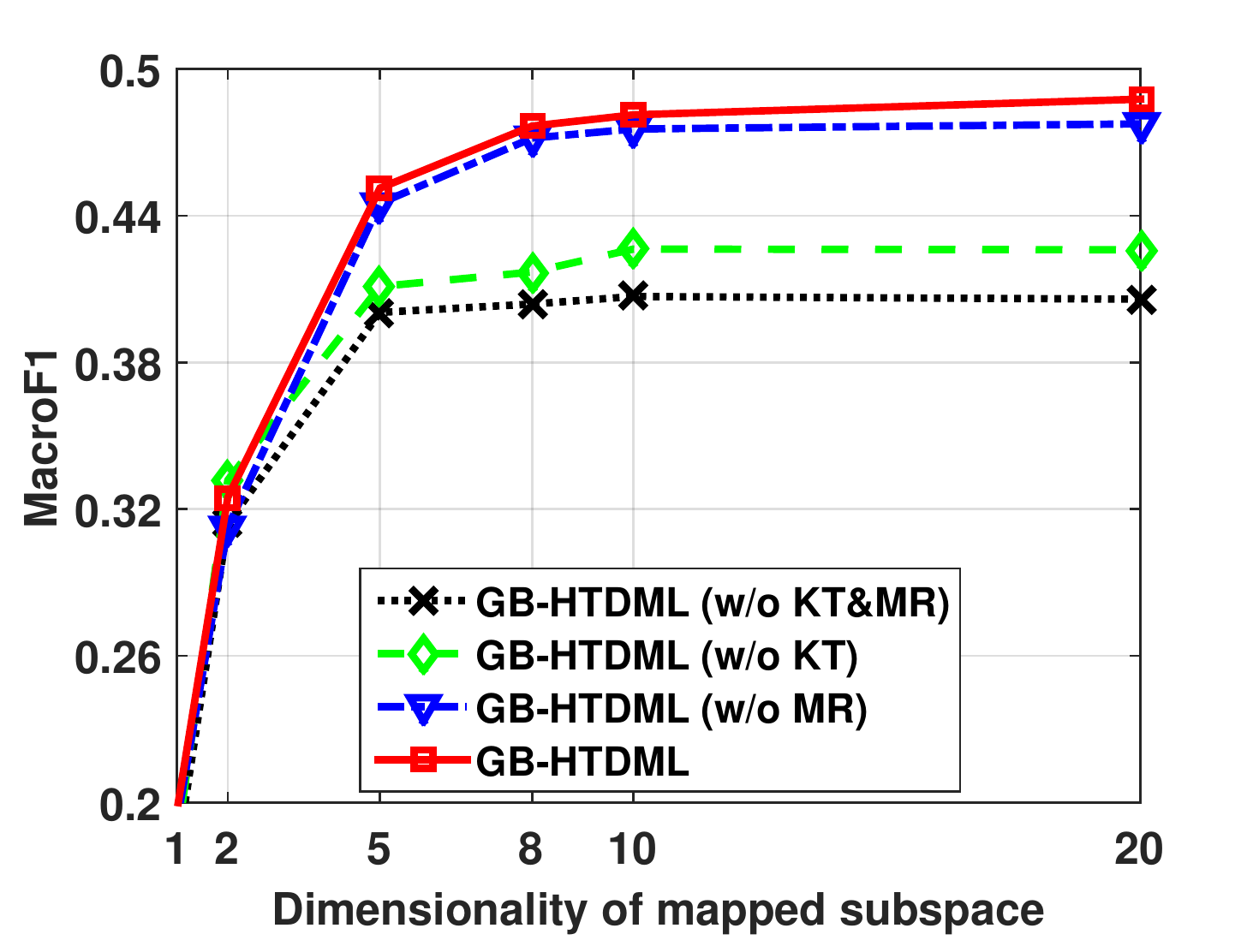}
}
\caption{A self-comparison of the proposed method on the Scene-15 dataset.}
\label{fig:Acc_MacF1_vs_Dim_SCE_GIST_Non_SC}
\end{figure}

To see how the two regularization terms affect the performance, we conduct a self-comparison of the proposed method. In particular, we compare GB-HTDML with the following three sub-models:
\begin{itemize}
  \item \textbf{GB-HTDML (w/o KT\&MR):} learning the target metric by only utilizing the labeled data in the target domain.
  \item \textbf{GB-HTDML (w/o KT):} learning the target metric using both the labeled and unlabeled data without knowledge transfer. This amounts to perform semi-supervised metric learning in the target domain.
  \item \textbf{GB-HTDML (w/o MR):} perform transfer metric learning without preserving topology using manifold regularization in the target domain.
\end{itemize}
The results are shown in Fig. \ref{fig:Acc_MacF1_vs_Dim_SCE_GIST_Non_SC}. From the results, we can see that: 1) preserving topology in the target domain can improve the performance of DML, but the improvements are not that large as transferring knowledge from the source domain. This demonstrates the superiority of the source domain knowledge and the effectiveness of the proposed transfer strategy; 2) the improvements achieved by topology preservation become small after the knowledge transfer. This may be because the unlabeled data are also leveraged for knowledge fragments transfer. Some unlabeled information has been utilized and the significance of the unlabeled data decreases.

\begin{figure}
\centering
\subfloat{\includegraphics[width=0.6\columnwidth]{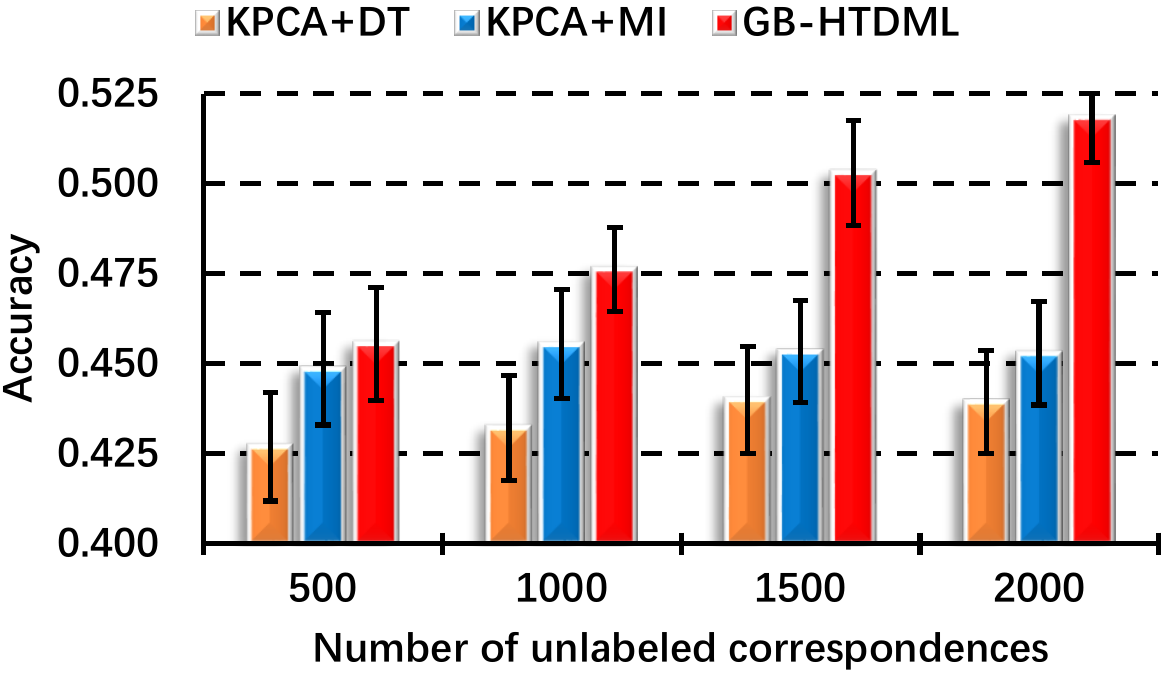}
}
\hfil
\subfloat{\includegraphics[width=0.6\columnwidth]{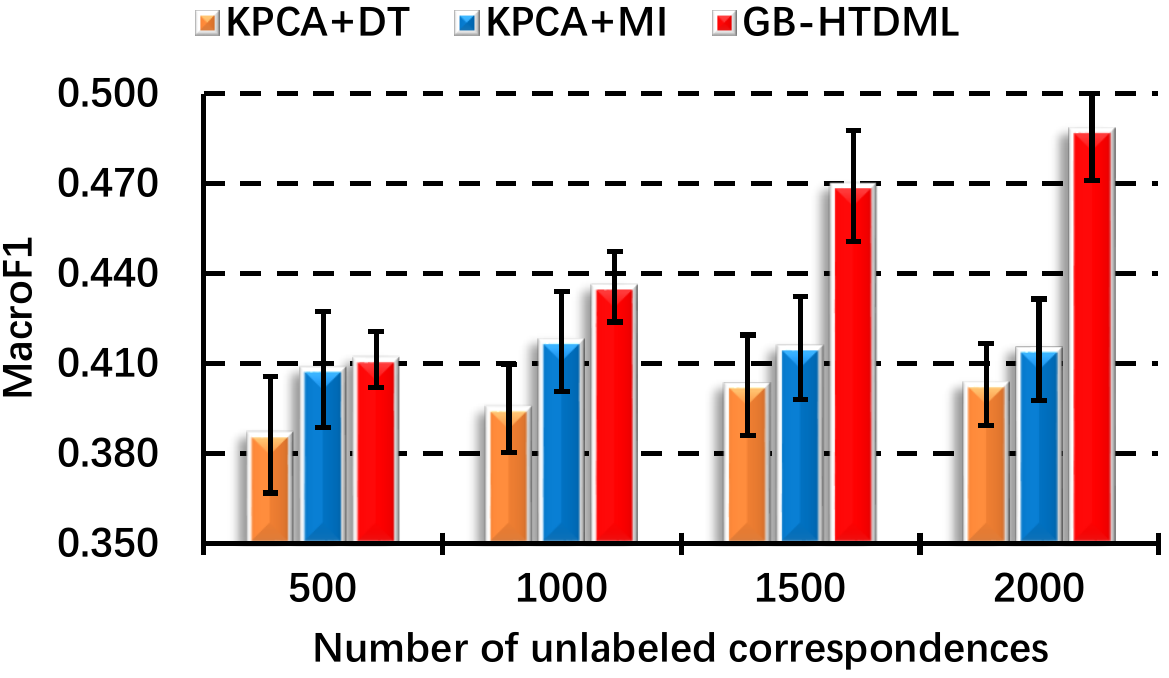}
}
\caption{Classification accuracies and macroF1 scores vs. number of unlabeled correspondences on the Scene-15 dataset.}
\label{fig:Acc_MacF1_vs_Corr}
\end{figure}

\subsubsection{An investigation of varying number of unlabeled corresponding pairs}
All DT, MI and the proposed HTDML make use of the unlabeled corresponding data between the source and target domain. In this set of experiments, we investigate how the number of correspondences affect their performance. We vary the number from $500$ to $2000$, and the experimental results are shown in Fig. \ref{fig:Acc_MacF1_vs_Corr}. It can be seen from the results that: 1) for DT and MI, there are only small improvements when the number of correspondences increases. The performance of DT and MI reach their peaks around $1500$ and $1000$ respectively. Whereas for the proposed GB-HTDML, the performance improves significantly with an increasing number of correspondences. This demonstrates that the KPCA preprocess cannot effectively capture the nonlinearity of the data distribution, which is properly discovered by our GB-HTDML; 2) the improvements of GB-HTDML decrease when the amounts of corresponding data are more than $1500$. This is because there is redundancy in the set of knowledge fragments, and the amount of new knowledge brought by including more correspondences becomes small when the size of the set is large enough.

\subsection{Object recognition}
In this set of experiments, we verify the proposed method in object recognition on a natural image dataset NUS-WIDE (NUS) \cite{TS-Chua-et-al-CIVR-2009}. The dataset contains $269,648$ images, and we conduct experiments on a subset that consists of $16,519$ images belonging to $12$ animal concepts. Half of the images are used for training and the rest for test. In this dataset, we choose the $1000$-D tag feature as source domain, and the $500$-D bag of SIFT \cite{DG-Lowe-IJCV-2004} visual words (BOVW) as the target domain. We vary the number of labeled instances for each concept in the set $\{20, 50, 100 \}$. This set of experiments is to see how much the easily interpretable text feature can guide the metric learning of visual feature, which is often harder to interpret \cite{GJ-Qi-et-al-SDM-2012}. The number of unlabeled correspondences is $5000$. For all methods, the dimensions of the tag and BOVW representations after the KPCA (or PCA in the proposed GB-HTDML) preprocess are both $100$.


\begin{table*}[!t]
\renewcommand{\arraystretch}{1.3}
\caption{Recognition accuracies and macroF1 scores of the compared methods on the NUS animal subset. In each domain, the number of labeled instances for each concept varies from $20$ to $100$.}
\label{tab:Acc_MacF1_Bst_NUS_Non}
\centering
\begin{tabular}{c||c|c|c||c|c|c}
\hline
Methods & \multicolumn{3}{c||}{Accuracy} & \multicolumn{3}{c}{MacroF1} \\
\hline
Methods & 20 & 50 & 100 & 20 & 50 & 100 \\
\hline
KPCA+EU & 0.195$\pm$0.010 & 0.223$\pm$0.010 & 0.245$\pm$0.003 & 0.220$\pm$0.007 & 0.248$\pm$0.009 & 0.266$\pm$0.003 \\
\hline
KPCA+LMNN & 0.199$\pm$0.014 & 0.253$\pm$0.006 & 0.268$\pm$0.004 & 0.225$\pm$0.008 & 0.267$\pm$0.008 & 0.280$\pm$0.004 \\
KPCA+ITML & 0.207$\pm$0.017 & 0.233$\pm$0.009 & 0.255$\pm$0.005 & 0.227$\pm$0.008 & 0.255$\pm$0.008 & 0.273$\pm$0.006 \\
\hline
KPCA+MTDA & 0.218$\pm$0.013 & 0.244$\pm$0.012 & 0.262$\pm$0.009 & 0.231$\pm$0.010 & 0.263$\pm$0.008 & 0.278$\pm$0.008 \\
KPCA+DAMA & 0.204$\pm$0.017 & 0.234$\pm$0.010 & 0.258$\pm$0.004 & 0.226$\pm$0.011 & 0.258$\pm$0.008 & 0.277$\pm$0.004 \\
\hline
KPCA+DT & 0.205$\pm$0.008 & 0.231$\pm$0.008 & 0.252$\pm$0.004 & 0.221$\pm$0.008 & 0.254$\pm$0.005 & 0.269$\pm$0.004 \\
KPCA+MI & 0.226$\pm$0.010 & 0.252$\pm$0.012 & 0.270$\pm$0.004 & 0.238$\pm$0.006 & 0.262$\pm$0.010 & 0.283$\pm$0.007 \\
\hline
GB-HTDML & \textbf{0.270$\pm$0.013} & \textbf{0.292$\pm$0.018} & \textbf{0.317$\pm$0.009} & \textbf{0.275$\pm$0.010} & \textbf{0.283$\pm$0.013} & \textbf{0.300$\pm$0.007} \\
\hline
\end{tabular}
\end{table*}

This set of experiments is more challenging than the scene categorization since there is semantic gap between the text and visual features \cite{GJ-Qi-et-al-SDM-2012}. From the results shown in Table \ref{tab:Acc_MacF1_Bst_NUS_Non}, we can see that: 1) the improvements of the single domain DML algorithms over the EU baseline are not that large as in scene categorization when the number of labeled data is small (e.g., $20$). This indicates that the effectiveness of the KPCA preprocess drops in this application; 2) DAMA is only comparable to the single domain DML algorithms. This may be led by the semantic gap. MTDA outperforms DAMA since the former learns an additional hidden layer. The transfer is conducted between higher level patterns and thus the semantic gap is reduced; 3) MI is superior to them when the labeled data are scarce. This mainly benefits from the large amounts of unlabeled correspondences and the manifold structure exploited in the source domain; 4) the proposed GB-HTDML outperforms all other approaches, but the improvements decrease when more labeled data are provided. This is because most of other approaches can achieve satisfactory performance when the labeled data are abundant. Nevertheless, the improvement is still significant when the number of labeled sample for each concept is large (e.g., $100$), since the nonlinear structure of the data distribution is appropriately explored in our method.


\subsection{Image retrieval}
In this section, we apply the proposed GB-HTDML method to image retrieval. The experiments are conducted on the popular Caltech-101 \cite{FF-Li-et-al-CVPRw-2004} dataset. The Caltech-101 dataset consists of $101$ object classes, and the number of images is $8,677$. Half of the dataset is used for training and the rest is for test. In the training set, we randomly select $2,000$ samples as the unlabeled data, and the remained images are labeled. The ``expensive'' CNN \cite{K-Chatfield-et-al-arXiv-2014} and ``cheap'' LBP \cite{T-Ojala-et-al-TPAMI-2002} features are utilized for data representation in the source and target domain respectively. Their features are provided by \cite{DX-Dai-et-al-CVPR-2015}, and we preprocess them using KPCA to exploit nonlinearity in the compared methods except the proposed GB-HTDML, where PCA is applied to reduce running time. The resulting dimensions are $100$ and $50$ for CNN and LBP respectively for all methods. Following \cite{DX-Dai-et-al-CVPR-2015}, we adopt mean average precision (MAP) as the evaluation criterion.

\begin{figure}
\centering
\includegraphics[width=0.6\columnwidth]{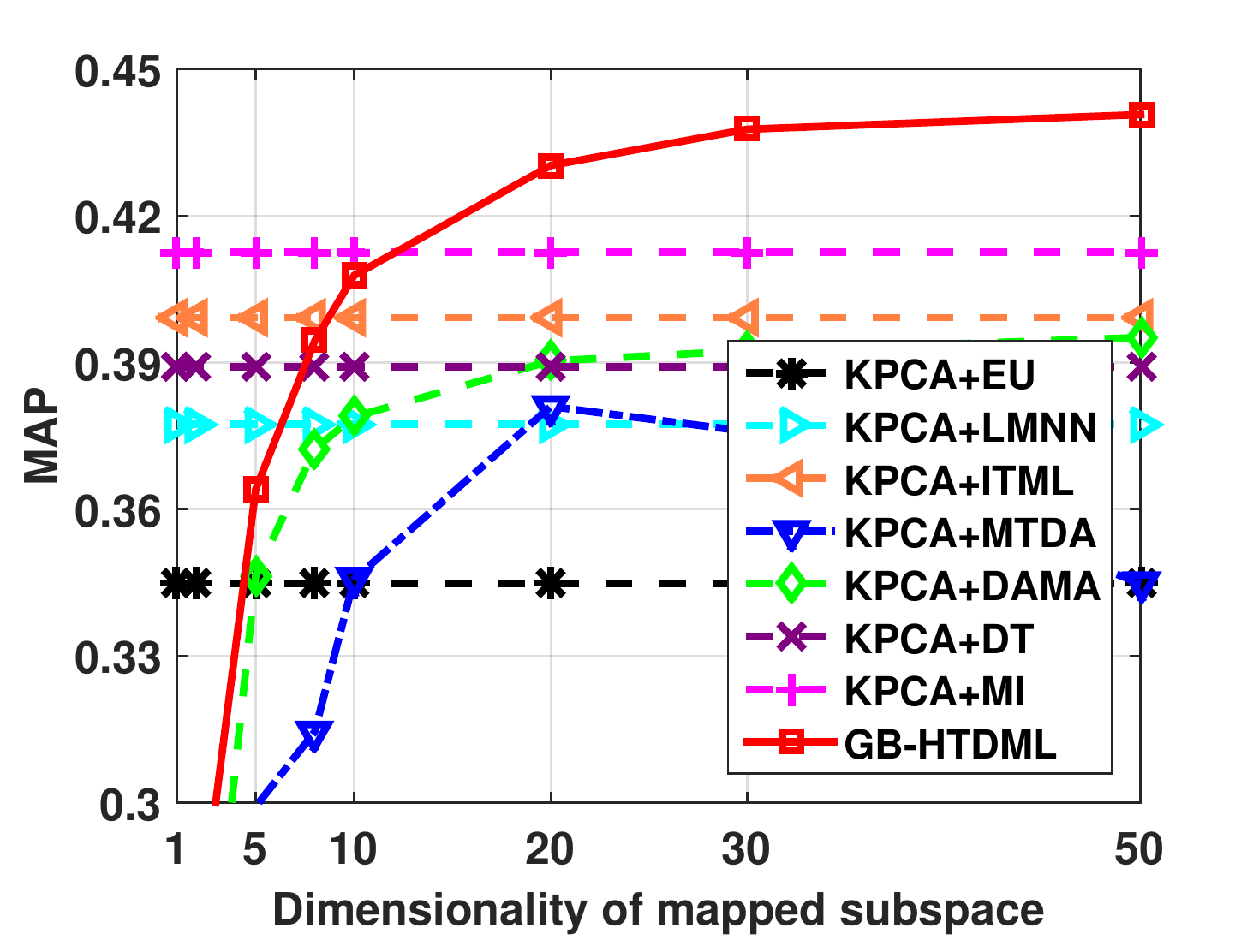}
\caption{Mean average precision (MAP) scores vs. dimensionality of the mapped subspace on the Caltech-101 dataset.}
\label{fig:Map_vs_Dim_Caltech_LBP_Non}
\end{figure}

The experimental results are shown in Fig. \ref{fig:Map_vs_Dim_Caltech_LBP_Non}. It can be observed from the results that: 1) the heterogeneous transfer learning (HTL) approaches DAMA and MTDA are only comparable to the single domain DML algorithms ITML and LMNN respectively. This may be because in this set of experiments, the label information is sufficient for the single domain DML algorithms to achieve satisfactory performance; 2) By making use of the unlabeled corresponding data, MI and the proposed GB-HTDML obtain much better performance than the other methods. This demonstrates the effectiveness of utilizing unlabeled correspondences for knowledge transfer. The proposed GB-HTDML is superior to MI since both the label information and unlabeled data are utilized, and the nonlinear structure of the data distribution is better exploited.

\subsection{Face verification}
We further verify the proposed method in the face verification application on the well-known labeled Faces in the Wild (LFW) \cite{GB-Huang-et-al-TR-2007} dataset. In the dataset, there are totally $13,233$ face images belonging to $5,749$ individuals. We conduct experiments under the ``unrestricted protocol''. That is, we only utilize the training data provided by LFW (no outside data). In the $10$-folds split of the dataset \cite{GB-Huang-et-al-TR-2007}, each fold is held-out for test in turn. In the remained $9$ folds, we randomly choose $5$ folds as labeled data, and the rest folds are unlabeled. CNN is still adopted for data representation in the source domain, and the baseline LBP features extracted by \cite{D-Chen-et-al-CVPR-2013} are utilized in the target domain. We reduce the dimension of the LBP feature to $400$ (suggested by \cite{D-Chen-et-al-CVPR-2013}) using PCA for the proposed GB-HTDML and KPCA for other approaches.

\begin{figure}
\centering
\includegraphics[width=0.6\columnwidth]{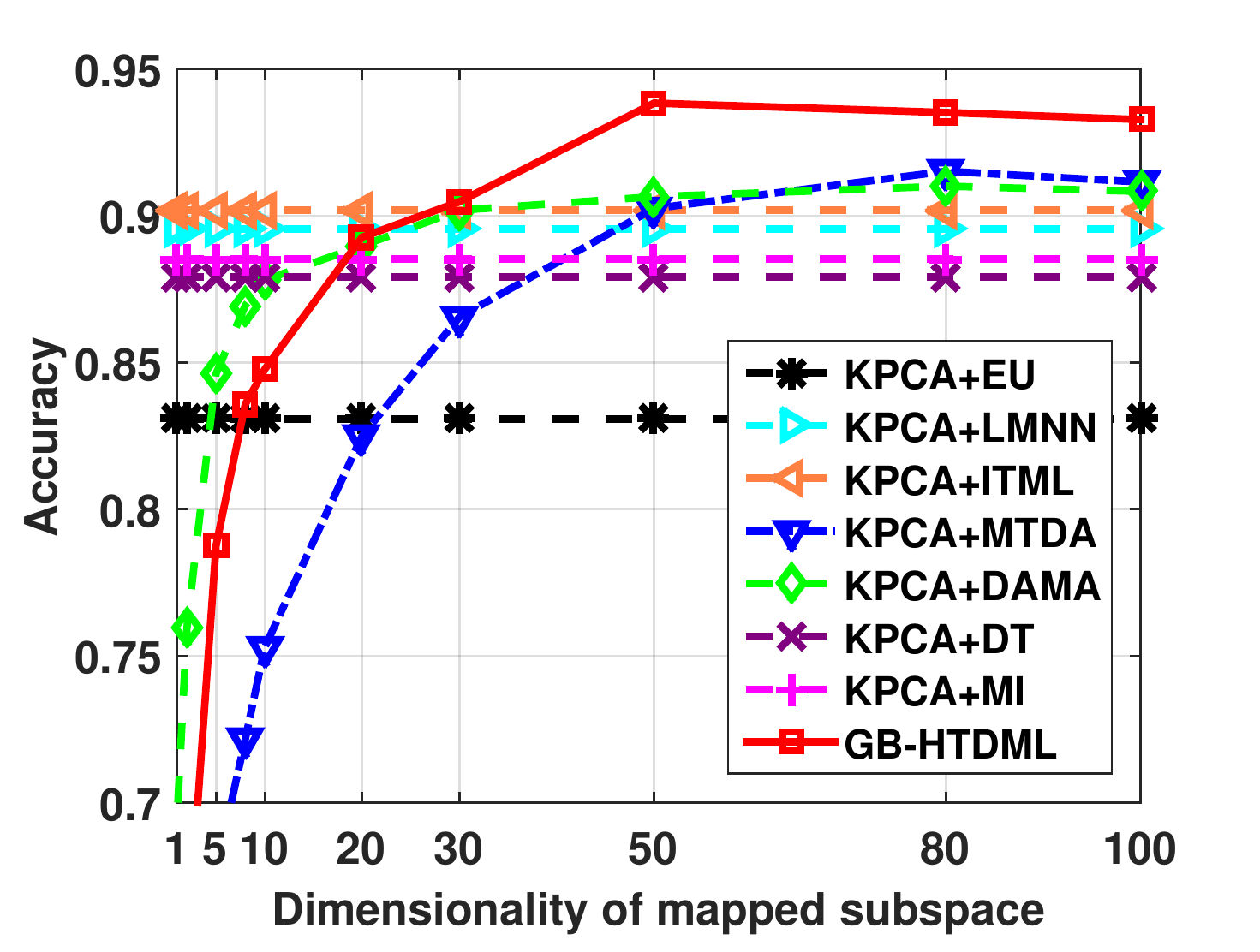}
\caption{Accuracies vs. dimensionality of the mapped subspace on the LFW dataset.}
\label{fig:Acc_vs_Dim_LFW_Non}
\end{figure}

The experimental results are shown in Fig. \ref{fig:Acc_vs_Dim_LFW_Non}, where we can see that: 1) both the single domain DML algorithms (LMNN and ITML) and HTL approaches (MTDA and DAMA) outperform the EU baseline significantly, and accuracies of the latter are a bit higher. This demonstrates the effectiveness of DML and advantages of knowledge transfer in this application; 2) LMNN and ITML are better than MI, while the proposed GB-HTDML is superior to all of them. This may be because MI only leverages unlabeled corresponding data for transfer, and the label information is more useful than the unlabeled correspondences in this application. By comparing this observation with that in the image retrieval application (where MI is quite competitive), we find that it is necessary to combine the label information and unlabeled correspondences to obtain a robust HTL model.


\section{Conclusion}\label{sec:Conclusion}
In this paper, we present a general heterogeneous transfer distance metric learning framework. The framework extracts a set of knowledge fragments from the source domain to help the metric learning in the target domain. Any existing distance metric learning (DML) algorithms can be adopted to learn the source knowledge fragments in an offline manner, and either linear or nonlinear metric can be learned for the target domain. Hence the proposed framework is general, flexible, and easy-to-use.

From the experimental evaluation on two popular applications, we mainly conclude that: 1) the performance of most of the current heterogeneous transfer learning (HTL) or metric transfer (imitation) approaches are unsatisfactory in the applications where data lie in a highly nonlinear feature space, or there is a semantic gap between the source and target domain. The KPCA preprocess can sometimes help the DML or HTL approaches to handle the data lie in a nonlinear feature space, but not always take effect; 2) by appropriately exploring the nonlinearity in both the source and target domains, we can obtain significant improvements over the KPCA counterpart.

It should be noted the ``cheap'' features are not restricted to the handcrafted features (such as LBP), but can also be some efficient CNN features, such as the ones extracted by some tiny version of the FaceNet model \cite{F-Schroff-et-al-CVPR-2015}. If the CNN feature extraction is more efficient but powerful than the handcrafted ``cheap'' feature (such as LBP) extraction, we can regard such CNN feature as the target domain representation, and using some more ``expensive'' CNN feature in the source domain to improve the performance of ``cheap'' CNN. In the future, we intend to adapt the proposed approach for other applications, such as large-scale video retrieval.

\ifCLASSOPTIONcaptionsoff
  \newpage
\fi



\bibliographystyle{IEEEtran}
\bibliography{./TPAMI-2017-04-0282}
%
%
%

%

%



\begin{IEEEbiography}[{\includegraphics[width=1in,height=1.25in,clip,keepaspectratio]{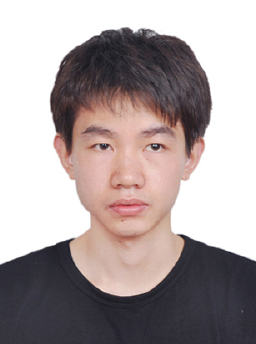}}]{Yong Luo}
received the B.E. degree in Computer Science from the Northwestern Polytechnical University, Xi'an, China, in 2009, and the D.Sc. degree in the School of Electronics Engineering and Computer Science, Peking University, Beijing, China, in 2014. He was a visiting student in the School of Computer Engineering, Nanyang Technological University, and the Faculty of Engineering and Information Technology, University of Technology Sydney. He is currently a Research Fellow with the School of Computer Science and Engineering, Nanyang Technological University. His research interests are primarily on machine learning with applications on multimedia classification and search. He has authored several scientific articles at top venues including IEEE T-NNLS, IEEE T-IP, IEEE T-KDE, IJCAI, and AAAI.
\end{IEEEbiography}

\begin{IEEEbiography}[{\includegraphics[width=1in,height=1.25in,clip,keepaspectratio]{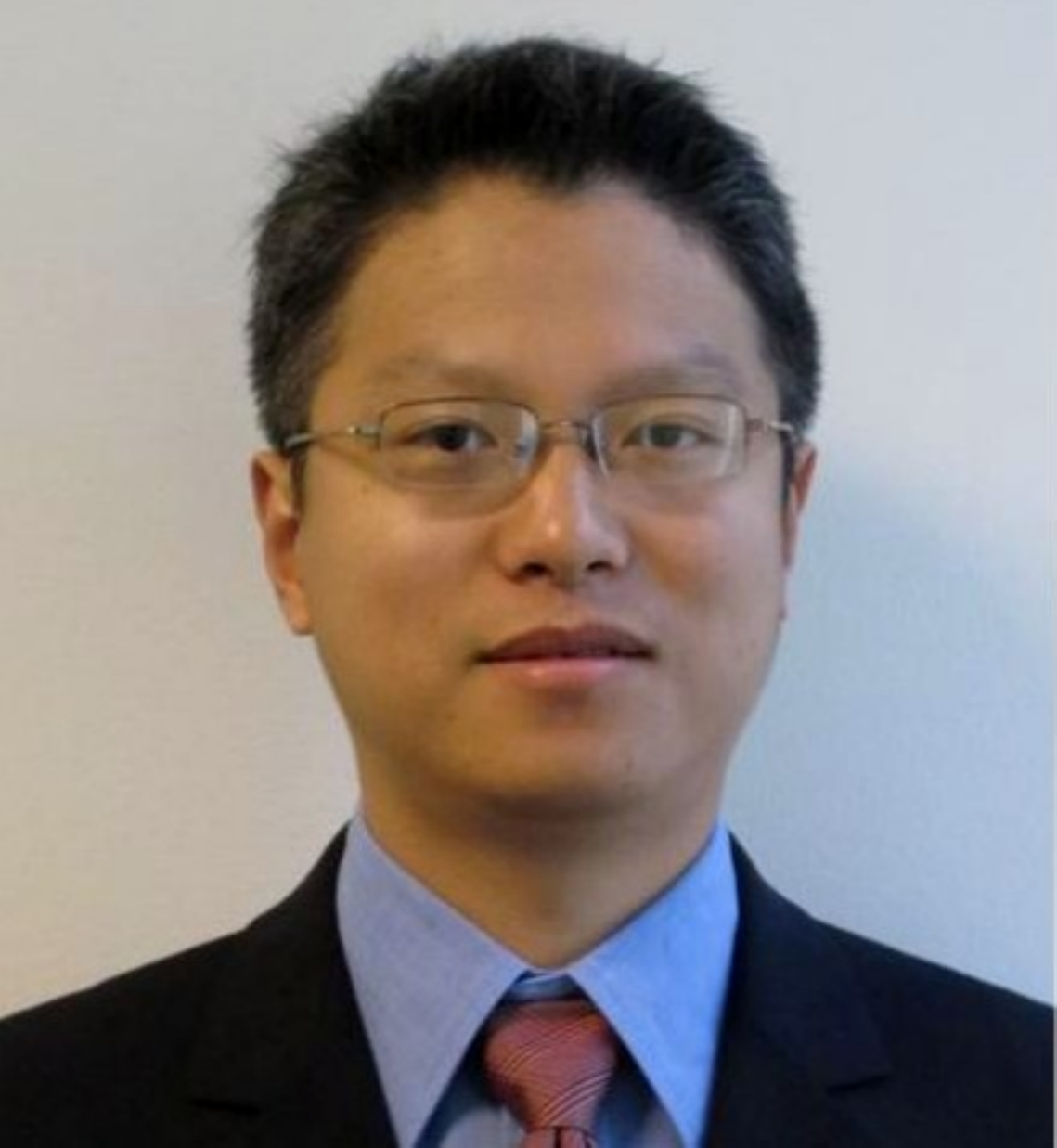}}]{Yonggang Wen}
(S'99-M'08-SM'14) received the Ph.D. degree in electrical engineering and computer science from the Massachusetts Institute of Technology, Cambridge, MA, USA, in 2008. He is currently an Assistant Professor with the School of Computer Engineering, Nanyang Technological University, Singapore. Previously, he worked at Cisco leading product development in content delivery network, which had a revenue impact of 3 billion U.S. dollars globally. He has published over 100 papers in top journals and prestigious conferences. His latest work in multi-screen cloud social TV has been featured by global media (more than 1600 news articles from over 29 countries) and recognized with the ASEAN ICT Award 2013 (Gold Medal) and the IEEE Globecom 2013 Best Paper Award. His research interests include cloud computing, green data center, big data analytics, multimedia network and mobile computing. Dr. Wen serves on the Editorial Boards of the IEEE TRANSACTIONS ON MULTIMEDIA, IEEE ACCESS, and Elsevier's Ad Hoc Networks.
\end{IEEEbiography}


\begin{IEEEbiography}[{\includegraphics[width=1in,height=1.25in,clip,keepaspectratio]{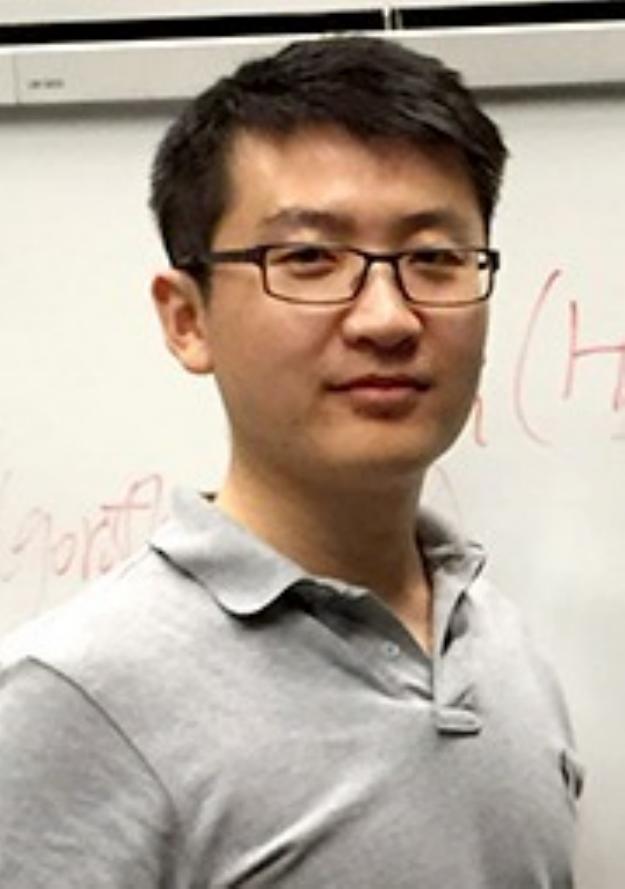}}]{Tongliang Liu}
received the B.E. degree in electronic engineering and information science from the University of Science and Technology of China, Hefei, China, in 2012, and the Ph.D. degree from the University of Technology Sydney, Sydney, Australia, in 2016. He was a visiting Ph.D. student with the Barcelona Graduate School of Economics and the Department of Economics, Pompeu Fabra University, for six months. He is currently a Lecturer with the School of Information Technologies and the Faculty of Engineering and Information Technologies, the University of Sydney. He has authored or co-authored over ten research papers, including the IEEE T-PAMI, T-NNLS, T-IP, NECO, ICML, KDD, IJCAI, and AAAI. His research interests include statistical learning theory, computer vision, and optimization. He received the Best Paper Award in the IEEE International Conference on Information Science and Technology 2014.
\end{IEEEbiography}

\begin{IEEEbiography}[{\includegraphics[width=1in,height=1.25in,clip,keepaspectratio]{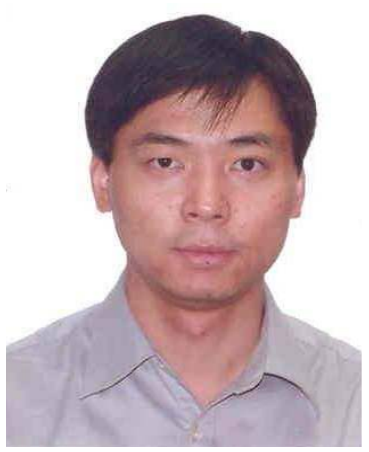}}]{Dacheng Tao}
(F'15) is currently a Professor of Computer Science with the School of Information Technologies and the Faculty of Engineering and Information Technologies, the University of Sydney. He mainly applies statistics and mathematics to artificial intelligence and data science. His research interests spread across computer vision, data science, image processing, machine learning, and video surveillance. His research results have expounded in one monograph and over 200 publications at prestigious journals and prominent conferences, such as the IEEE T-PAMI, T-NNLS, T-IP, JMLR, IJCV, NIPS, ICML, CVPR, ICCV, ECCV, AISTATS, ICDM, and ACM SIGKDD, with several best paper awards, such as the best theory/algorithm paper runner up award in the IEEE ICDM07, the best student paper award in the IEEE ICDM13, and the 2014 ICDM 10-year highest-impact paper award. He is a fellow of the IEEE, OSA, IAPR, and SPIE. He received the 2015 Australian Scopus-Eureka Prize, the 2015 ACS Gold Disruptor Award, and the 2015 UTS Vice-Chancellors Medal for Exceptional Research.
\end{IEEEbiography}




\end{document}